\documentclass[10pt,journal,compsoc]{IEEEtran}
\ifCLASSOPTIONcompsoc
  \usepackage[nocompress]{cite}
\else
  \usepackage{cite}
\fi
\ifCLASSINFOpdf
\else
\fi
\usepackage{graphicx}
\usepackage{CJK,CJKnumb}
\usepackage{bbm}
\usepackage{amsmath,bm}
\usepackage{color}
\usepackage{makecell}
\usepackage{caption}
\usepackage{multirow}
\usepackage[figuresright]{rotating}
\usepackage{array}
\usepackage{longtable, booktabs}
\usepackage[font=normalsize]{caption}

\begin{document}
%
\title{Scene Graph Generation: A Comprehensive Survey }
%
%
%
%

\author{Guangming~Zhu,~Liang~Zhang,~Youliang~Jiang,~Yixuan~Dang,~Haoran~Hou,~Peiyi~Shen*,~Mingtao~Feng,~Xia~Zhao*,~Qiguang~Miao,~Syed~Afaq~Ali~Shah~and~Mohammed~Bennamoun
\IEEEcompsocitemizethanks{\IEEEcompsocthanksitem G. Zhu, L. Zhang, Y. Jiang, Y. Dang, H. Hou, P. Shen, M. Feng and Q. Miao are with the School of Computer Science and Technology, Xidian University, 710071, Xian, China (E-mail: gmzhu@xidian.edu.cn; liang.zhang.cn@ieee.org; mqjyl2012@163.com; dyx4work@gmail.com; unse3ry@gmail.com; pyshen@xidian.edu.cn; mtfeng@xidian.edu.cn; qgmiao@xidian.edu.cn), * indicates the corresponding author.
\IEEEcompsocthanksitem X. Zhao is with the School of Arts and Sciences, National University of Defense Technology, Changsha, China (E-mail: zxmdi@163.com)
\IEEEcompsocthanksitem S. A. A. Shah is with the centre for AI and Machine Learning, Edith Cowan University, Australia (E-mail: afaq.shah@ecu.edu.au)
\IEEEcompsocthanksitem M. Bennamoun is with School of Computer Science and Software Engineering, The University of Western Australia, Australia (E-mail: mohammed.bennamoun@uwa.edu.au)}
\thanks{Manuscript received January XX, 2021. This work is supported by National Natural Science Foundation of China(62073252, 62072358), National Key R\&D Program of China under Grant No.2020YFF0304900, 2019YFB1311600, and Chinese Defense Advance Research Program (50912020105).}}

\IEEEtitleabstractindextext{%
\begin{abstract}
Deep learning techniques have led to remarkable breakthroughs in the field of generic object detection and have spawned a lot of scene-understanding tasks in recent years. Scene graph has been the focus of research because of its powerful semantic representation and applications to scene understanding. Scene Graph Generation (SGG) refers to the task of automatically mapping an image or a video into a semantic structural scene graph, which requires the correct labeling of detected objects and their relationships. Although this is a challenging task, the community has proposed a lot of SGG approaches and achieved good results. In this paper, we provide a comprehensive survey of recent achievements in this field brought about by deep learning techniques. We review 138 representative works, and systematically summarize existing methods of image-based SGG from the perspective of feature representation and refinement. We attempt to connect and systematize the existing visual relationship detection methods, to summarize, and interpret the mechanisms and the strategies of SGG in a comprehensive way. Finally, we finish this survey with deep discussions about current existing problems and future research directions. This survey will help readers to develop a better understanding of the current research status and ideas.
\end{abstract}

}

\maketitle

\IEEEdisplaynontitleabstractindextext

%
\IEEEpeerreviewmaketitle

\section{Introduction}

%
%
%
%


The ultimate goal of computer vision (CV) is to build intelligent systems, which can extract valuable information from digital images, videos, or other modalities as humans do. In the past decades, machine learning (ML) has significantly contributed to the progress of CV. Inspired by the ability of humans to interpret and understand visual scenes effortlessly, \emph{visual scene understanding} has long been advocated as the holy grail of CV and has already attracted much attention from the research community. 

\begin{figure*}[tp] 
	\centering 
     	\includegraphics[width=\textwidth]{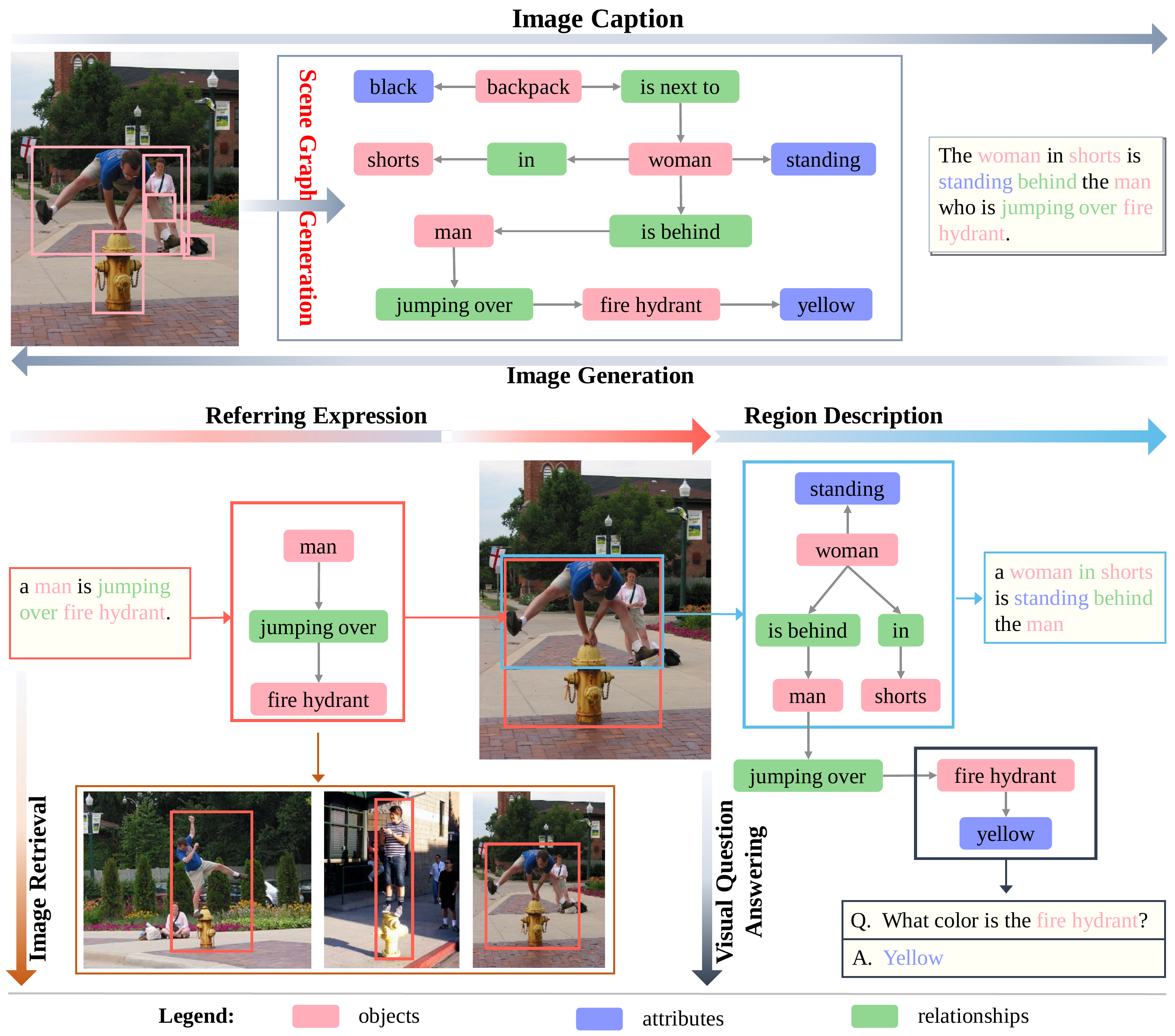} 
	\caption{\normalsize{A visual illustration of a scene graph structure and some applications. \bfseries Scene graph generation \mdseries models take an image as an input and generate a visually-grounded scene graph. \bfseries Image caption \mdseries can be generated from a scene graph directly. In contrast, \bfseries Image generation \mdseries inverts the process by generating realistic images from a given sentence or scene graph. The \bfseries Referring Expression (REF) \mdseries marks a region of the input image corresponding to the given expression, while the region and expression map the same subgraph of the scene graph. Scene graph-based \bfseries image retrieval \mdseries takes a query as an input, and regards the retrieval as a scene graph matching problem. For the \bfseries Visual Question Answering (VQA) \mdseries task, the answer can sometimes be found directly on the scene graph, even for the more complex visual reasoning, the scene graph is also helpful.}}
	\label{fig:scene-graph-structure}                          
\end{figure*}

Visual scene understanding includes numerous sub-tasks, which can be generally divided into two parts: recognition and application tasks. These recognition tasks can be described at several semantic levels. Most of the earlier works, which mainly concentrated on image classification, only assign a single label to an image, e.g., an image of a cat or a car, and go further in assigning multiple annotations without localizing where in the image each annotation belongs\cite{li2003automatic}. A large number of neural network models have emerged and even achieved near humanlike performance in image classification tasks\cite{he2016deep, szegedy2017inception, chollet2017xception, huang2017densely}. Furthermore, several other complex tasks, such as semantic segmentation at the pixel level, object detection and instance segmentation at the instance level, have suggested the decomposition of an image into foreground objects vs background clutter. The pixel-level tasks aim at classifying each pixel of an image (or several) into an instance, where each instance (or category) corresponds to a class\cite{taghanaki2020deep}. The instance-level tasks focus on the detection and recognition of individual objects in the given scene and delineating an object with a bounding box or a segmentation mask, respectively. A recently proposed approach named Panoptic Segmentation (PS) takes into account both per-pixel class and instance labels\cite{kirillov2019panoptic}. With the advancement of Deep Neural Networks (DNN), we have witnessed important breakthroughs in object-centric tasks and various commercialized applications based on existing state-of-the-art models\cite{ronneberger2015u, chen2018encoder, ren2015faster, redmon2018yolov3, chen2019hybrid}. However, scene understanding goes beyond the localization of objects. The higher-level tasks lay emphasis on exploring the rich semantic relationships between objects, as well as the interaction of objects with their surroundings, such as visual relationship detection (VRD)\cite{li2017vip, liang2018visual, zhang2019large, li2017scene} and human-object interaction (HOI)\cite{gkioxari2018detecting, qi2018learning, wang2019deep}. These tasks are equally significant and more challenging. To a certain extent, their development depends on the performance of individual instance recognition techniques. Meanwhile, the deeper semantic understanding of image content can also contribute to visual recognition tasks\cite{grzeszick2016zero, deng2019relation, liu2018structure, hu2018relation, krishna2018referring}. Divvala \emph{et al}.\cite{divvala2009empirical} investigated various forms of context models, which can improve the accuracy of object-centric recognition tasks. In the last few years, researchers have combined computer vision with natural language processing (NLP) and proposed a number of advanced research directions, such as image captioning, visual question answering (VQA), visual dialog and so on. These vision-and-language topics require a rich understanding of our visual world and offer various application scenarios of intelligent systems. 

Although rapid advances have been achieved in the scene understanding at all levels, there is still a long way to go. Overall perception and effective representation of information are still bottlenecks. As indicated by a series of previous works\cite{Johnson2015Image, bao2012semantic, armeni20193d}, building an efficient structured representation that captures comprehensive semantic knowledge is a crucial step towards a deeper understanding of visual scenes. Such representation can not only offer contextual cues for fundamental recognition challenges, but also provide a promising alternative to high-level intelligence vision tasks. \emph{Scene graph}, proposed by Johnson \emph{et al}.\cite{Johnson2015Image}, is a visually-grounded graph over the object instances in a specific scene, where the nodes correspond to object bounding boxes with their object categories, and the edges represent their pair-wise relationships.

Because of the structured abstraction and greater semantic representation capacity compared to image features, scene graph has the instinctive potential to tackle and improve other vision tasks. As shown in {Fig.\ref{fig:scene-graph-structure}}, a scene graph parses the image to a simple and meaningful structure and acts as a bridge between the visual scene and textual description. Many tasks that combine vision and language can be handled with scene graphs, including image captioning\cite{gao2018image, yang2019auto, kim2019dense}, visual question answering\cite{zhang2019empirical, li2019relation}, content-based image retrieval \cite{Johnson2015Image, schuster2015generating}, image generation\cite{johnson2018image, mittal2019interactive} and referring expression comprehension\cite{yang2019cross}. Some tasks take an image as an input and parse it into a scene graph, and then generate a reasonable text as output. Other tasks invert the process by extracting scene graphs from the text description and then generate realistic images or retrieve the corresponding visual scene. 

Xu \emph{et al}.\cite{xu2020survey} have produced a thorough survey on scene graph generation, which analyses the SGG methods based on five typical models (CRF, TransE, CNN, RNN/LSTM and GNN) and also includes a discussion of important contributions by prior knowledge. Moreover, a detailed investigation of the main applications of scene graphs was also provided. 
\textcolor{black}{The current survey focuses on visual relationship detection of SGG, and our survey's organization is based on feature representation and refinement. Specifically, we first provide a comprehensive and systematic review of 2D SGG. In addition to multimodal features, prior information and commonsense knowledge to help overcome the long-tailed distribution and the large intra-class diversity problems, is also provided. To refine the local features and fuse the contextual information for high-quality relationship prediction, we analyze some mechanisms, such as message passing, attention, and visual translation embedding. In addition to 2D SGG, spatio-temporal and 3D SGG are also examined. Further, a detailed discussion of the most common datasets is provided together with performance evaluation measures. Finally, a comprehensive and systematic review of the most recent research on the generation of scene graphs is presented.}
We provide a survey of 138 papers on SGG$\footnote{We provide a curated list of scene graph generation methods, publicly accessible at https://github.com/mqjyl/awesome-scene-graph}$, which have appeared since 2016 in the leading computer vision, pattern recognition, and machine learning conferences and journals. Our goal is to help the reader study and understand this research topic, which has gained a significant momentum in the past few years. The main contributions of this article are as follows:

\begin{enumerate}
	
	\item \textcolor{black}{A comprehensive review of 138 papers on scene graph generation is presented, covering nearly all of the current literature on this topic.} 
	
	\item \textcolor{black}{A systematic analysis of 2D scene graph generation is presented, focusing on feature representation and refinement. The long-tail distribution problem and the large intra-class diversity problem are addressed from the perspectives of fusing prior information and commonsense knowledge, as well as refining features through message passing, attention, and visual translation embedding. }
	
	\item \textcolor{black}{A review of typical datasets for 2D, spatio-temporal and 3D scene graph generation is presented, along with an analysis of the performance evaluation of the corresponding methods on these datasets.}
	
\end{enumerate}      

The rest of this paper is organized as follows; Section 2 gives the definition of a scene graph, thoroughly analyses the characteristics of visual relationships and the structure of a scene graph. Section 3 surveys scene graph generation methods. Section 4 summarizes almost all currently published datasets. Section 5 compares and discusses the performance of some key methods on the most commonly used datasets. Finally, Section 6 summarizes open problems in the current research and discusses potential future research directions. Section 7 concludes the paper.

\section{Scene Graph}
\label{sec:Scene_Graph}

A scene graph is a structural representation, which can capture detailed semantics by explicitly modeling objects (``man", ``fire hydrant", ``shorts"), attributes of objects (``fire hydrant is yellow"), and relations between paired objects (``man jumping over fire hydrant"), as shown in {Fig.\ref{fig:scene-graph-structure}}. The fundamental elements of a scene graph are objects, attributes and relations. \textcolor{black}{Subjects/objects are the core building blocks of an image and they can be located with bounding boxes. Each object can have zero or more attributes, such as color (e.g., yellow), state (e.g., standing), material (e.g., wooden), etc. Relations can be actions (e.g., ``jump over"), spatial (e.g., ``is behind"), descriptive verbs (e.g., wear), prepositions (e.g. ``with"), comparatives (e.g., ``taller than"), prepositional phrases (e.g., ``drive on"), etc\cite{liang2017deep, krishna2017visual, kuznetsova2018open, lu2016visual}}. In short, a scene graph is a set of \emph{visual relationship triplets} in the form of $\left\langle subject, relation, object\right\rangle$ or $\left\langle object, is, attribute\right\rangle$. The latter is also considered as a relationship triplet (using the ``is" relation for uniformity\cite{kuznetsova2018open, zhang2018interpretable}).

In this survey paper, we mainly focus on the triplet description of a static scene. Given a visual scene $S \in \mathcal{S}$\cite{qi2019attentive}, such as an image or a 3D mesh, its scene graph is a set of visual triplets $\mathcal{R}_S \subseteq \mathcal{O}_S \times \mathcal{P}_S \times (\mathcal{O}_S \cup \mathcal{A}_S)$, where $\mathcal{O}_S$ is the object set, $\mathcal{A}_S$ is the attribute set and $\mathcal{P}_S$ is the relation set including ``is" relation $p_{S, is}$ where there is only one object involved. Each object $o_{S, k} \in \mathcal{O}_S$ has a semantic label $l_{S, k} \in \mathcal{O}_L$ ($\mathcal{O}_L$ is the semantic label set) and grounded with a bounding box (BB) $b_{S, k}$ in scene $S$, where $k \in \{1,\ldots,|\mathcal{O}_S|\}$. Each relation $p_{S, i\rightarrow j} \in \mathcal{P}_S \subseteq \mathcal{P}$ is the core form of a visual relationship triplet $r_{S, i\rightarrow j} = \left\langle o_{S, i}, p_{S, i\rightarrow j}, o_{S, j} \right\rangle \in \mathcal{R}_S$ and $i \neq j$, where the third element $o_{S, j}$ could be an attribute $a_{S, j} \subseteq \mathcal{A}_S$ if $p_{S, i\rightarrow j}$ is the $p_{S, is}$. As the relationship is one-way, we expresse $r_{S, i\rightarrow j}$ as $\left\langle s_{S, i}, p_{S, i\rightarrow j}, o_{S, j} \right\rangle$ to maintain semantic accuracy where $s_{S, i}, o_{S, j} \in \mathcal{O}_S$, $s_{S, i}$ is \emph{subject} and $o_{S, j}$ is \emph{object}.

From the point of view of graph theory, a scene graph is a directed graph with three types of nodes: object, attribute, and relation. However, for the convenience of semantic expression, a node of a scene graph is seen as an object with all its attributes, while the relation is called an edge. A subgraph can be formed with an object, which is made up of all the related visual triplets of the object. Therefore, the subgraph contains all the adjacent nodes of the object, and these adjacent nodes directly reflect the context information of the object. From the top-down view, a scene graph can be broken down into several subgraphs, a subgraph can be splitted into several triplets, and a triplet can be splitted into individual objects with their attributes and relations. Accordingly, we can find a region in the scene corresponding to the substructure that is a subgraph, a triplet, or an object. \textcolor{black}{Clearly, a perfectly-generated scene graph corresponding to a given scene should be structurally unique. The process of generating a scene graph should be objective and should only be dependent on the scene. Scene graphs should serve as an objective semantic representation of the state of the scene. The SGG process should not be affected by who labelled the data, on how it was assigned objects and predicate categories, or on the performance of the SGG model used. Although, in reality, not all annotators who label the data give produce the exact same visual relationship for each triplet, and the methods that generate scene graphs do not always predict the correct relationships.} The uniqueness supports the argument that the use of a scene graph as a replacement for a visual scene at the language level is reasonable. 

Compared with scene graphs, the well-known \emph{knowledge graph} is represented as multi-relational data with enormous fact triplets in the form of (\emph{head entity type}, \emph{relation}, \emph{tail entity type})\cite{ji2020survey, wan2018representation}. Here, we have to emphasize that the visual relationships in a scene graph are different from those in social networks and knowledge bases. In the case of vision, images and visual relationships are incidental and are not intentionally constructed. Especially, visual relationships are usually image-specific because they only depend on the content of the particular image in which they appear. Although a scene graph is generated from a textual description in some language-to-vision tasks, such as image generation, the relationships in a scene graph are always situation-specific. Each of them has the corresponding visual feature in the output image. Objects in scenes are not independent and tend to cluster. Sadeghi \emph{et al}.\cite{sadeghi2011recognition} coined the term \emph{visual phrases} to introduce composite intermediates between objects and scenes. Visual phrases, which integrate linguistic representations of relationship triplets encode the interactions between objects and scenes. 

\textcolor{black}{A two-dimensional (2D) image is a projection of a three-dimensional (3D) world from a particular perspective. Because of the visual shade and dimensionality reduction caused by the projection of 3D to 2D, 2D images may have incomplete or ambiguous information about the 3D scene, leading to an imperfect representation of 2D scene graphs. As opposed to a 2D scene graph, a 3D scene graph prevents spatial relationship ambiguities between object pairs caused by different viewpoints. } The relationships described above are static and instantaneous because the information is grounded in an image or a 3D mesh that can only capture a specific moment or a certain scene. On the other hand, with videos, a visual relationship is not instantaneous, but varies with time. A digital video consists of a series of images called frames, which means relations span over multiple frames and have different durations. Visual relationships in a video can construct a \textit{Spatio-Temporal Scene Graph}, which includes entity nodes of the neighborhood in the time and space dimensions. 

\textcolor{black}{The scope of our survey therefore extends beyond the generation of 2D scene graphs to include 3D and spatiotemporal scene graphs as well.}

\section{Scene Graph Generation}

\textcolor{black}{The goal of scene graph generation is to parse an image or a sequence of images in order to generate a structured representation, to bridge the gap between visual and semantic perception, and ultimately to achieve a complete understanding of visual scenes. However, it is difficult to generate an accurate and complete scene graph. Generating a scene graph is generally a bottom-up process in which entities are grouped into triplets and these triplets are connected to form the entire scene graph. Evidently, the essence of the task is to detect the visual relationships, i.e. $\left\langle \emph{subject, relation, object}\right\rangle$ triplets, abbreviated as $\left\langle \emph{s, r, o}\right\rangle$. Methods, which are used to connect the detected visual relationships to form a scene graph, do not fall in the scope of this survey. This paper focuses on reviewing methods for visual relationship detection.}

Visual Relationship Detection has attracted the attention of the research community since the pioneering work by Lu \emph{et al}\cite{lu2016visual}, and the release of the ground-breaking large-scale scene graph dataset Visual Genome (VG) by Krishna \emph{et al}\cite{krishna2017visual}. Given a visual scene $S$ and its scene graph $\mathcal{T}_S$\cite{chen2019knowledge, qi2019attentive}: 

\begin{itemize}
\item $\mathcal{B}_S = \{b_{S,1}, \ldots, b_{S,n}\}$ is the region candidate set, with element $b_{S,i}$ denoting the bounding box of the \emph{i}-th candidate object.
\item $\mathcal{O}_S = \{o_{S,1},\ldots,o_{S,n}\}$ is the object set, with element $o_{S,i}$ denoting the corresponding class label of the object $b_{S,i}$.
\item $\mathcal{A}_S = \{a_{S,o_1,1}, \ldots, a_{S,o_1,k_1}, \ldots, a_{S,o_2,1}, \ldots, a_{S,o_2,k_2}, \\ \ldots, a_{S,o_n,1}, \ldots, a_{S,o_n,k_n},\}$ is the attribute set, with element $a_{S,o_i,j}$ denoting the \emph{j}-th attribute of the \emph{i}-th object, where $k_i \ge 0$ and $ j \in \{1,\ldots,k_i\}$.
\item $\mathcal{R}_S = \{r_{S,1\rightarrow 2}, r_{S,1\rightarrow 3},\ldots,r_{S, n\rightarrow n-1}\}$ is the relation set, with element $r_{S,i\rightarrow j}$ corresponding to a visual triple $t_{S, i\rightarrow j} = \left\langle s_{S, i}, r_{S, i\rightarrow j}, o_{S, j}\right\rangle$, where $s_{S, i}$ and $o_{S, j}$ denote the subject and object respectively. This set also includes ``is" relation where there is only one object invovled. 
\end{itemize}

When attributes detection and relationships prediction are considered as two independent processes, we can decompose the probability distribution of the scene graph $p(\mathcal{T}_S | S)$ into four components similar to \cite{chen2019knowledge}:
\begin{equation}\label{eq:SGG_probability}
\begin{aligned}
p(\mathcal{T}_S | S) = & p(\mathcal{B}_S | S)p(\mathcal{O}_S |\mathcal{B}_S, S) \\ & (p(\mathcal{A}_S |\mathcal{O}_S, \mathcal{B}_S, S)p(\mathcal{R}_S |\mathcal{O}_S, \mathcal{B}_S, S))
\end{aligned}
\end{equation}

In the equation, the bounding box component $p(\mathcal{B}_S | S)$ generates a set of candidate regions that cover most of the crucial objects directly from the input image. The object component $p(\mathcal{O}_S |\mathcal{B}_S, S)$ predicts the class label of the object in the bounding box. Both steps are identical to those used in two-stage target detection methods, and can be implemented by the widely used Faster RCNN detector\cite{ren2015faster}. Conditioned on the predicted labels, the attribute component $p(\mathcal{A}_S |\mathcal{O}_S, \mathcal{B}_S, S)$ infers all possible attributes of each object, while the relationship component $p(\mathcal{R}_S |\mathcal{O}_S, \mathcal{B}_S, S)$ infers the relationship of each object pair\cite{chen2019knowledge}. When all visual triplets are collected, a scene graph can then be constructed. Since attribute detection is generally regarded as an independent research topic, visual relationship detection and scene graph generation are often regarded as the same task. Then, the probability of a scene graph $\mathcal{T}_S$ can be decomposed into three factors:

\begin{equation}\label{eq:SGG_probability_simple}
\begin{aligned}
p(\mathcal{T}_S | S) = p(\mathcal{B}_S | S) p(\mathcal{O}_S |\mathcal{B}_S, S) p(\mathcal{R}_S |\mathcal{O}_S, \mathcal{B}_S, S)
\end{aligned}
\end{equation}

The following section provides a detailed review of more than a hundred deep learning-based methods proposed until 2020 on visual relationship detection and scene graph generation. \textcolor{black}{In view of the fact that 2D SGG has been published much more than 3D or spatio-temporal SGG, a comprehensive overview of the methods for 2D SGG is first provided.  This is followed by a review of the 3D and spatiotemporal SGG methods in order to ensure completeness and breadth of the survey.}

\textcolor{black}{Note: We use ``relationship" or a ``triplet" to refer to the tuple of $\left\langle \emph{subject, relation, object}\right\rangle$ in this paper, and “relation” or a ``predicate" to refer to a \emph{relation} element. }

\subsection{2D Scene Graph Generation}

\begin{figure*}[tp] 
	\centering 
	\includegraphics[width=\textwidth]{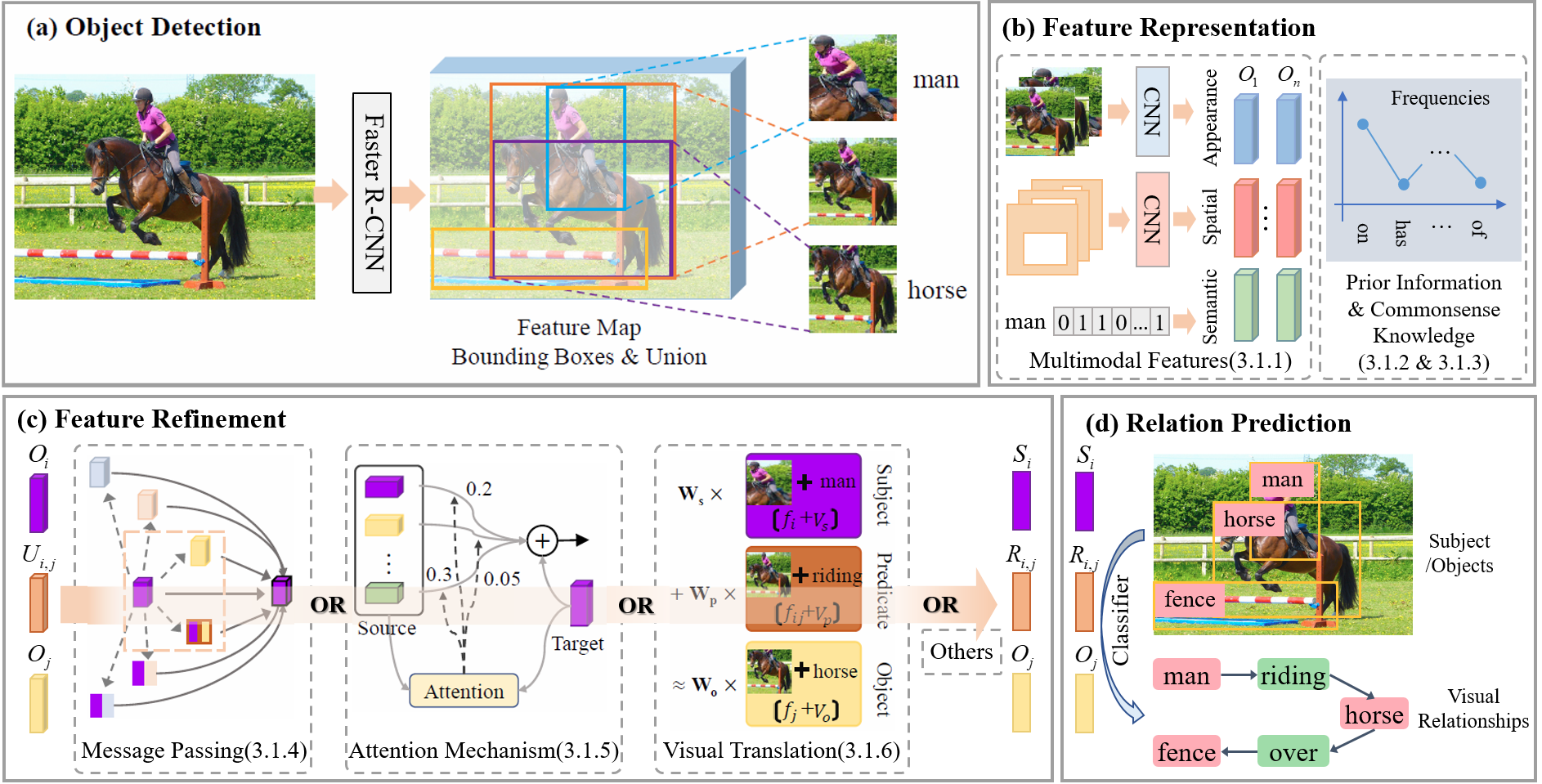} 
	\caption{\textcolor{black}{An overview of 2D general scene graph generation framework. Firstly, off-the-shelf object detectors are used to detect subjects, objects and predicate ROIs. Then, different kinds of methods are used in the stages of \textbf{(b) Feature Representation} and \textbf{(c) Feature Refinement} to improve the final \textbf{(d) Relation Prediction} for high-quality visual relationship detection. This survey focuses on the methods of feature representation and refinement.}}
	\label{fig:scene-graph-generation-process}  
\end{figure*}

Scene graphs can be generated in two different ways\cite{li2018factorizable}. The mainstream approach uses a two-step pipeline that detects objects first and then solves a classification task to determine the relationship between each pair of objects. The other approach involves jointly inferring the objects and their relationships based on the object region proposals. Both of the above approaches need to first detect all existing objects or proposed objects in the image, and group them into pairs and use the features of their union area (called relation features), as the basic representation for the predicate inference. In this section, we focus on the two-step approach, and {Fig.\ref{fig:scene-graph-generation-process}} illustrates the general framework for creating 2D scene graphs. Given an image, \textcolor{black}{a scene graph generation method first generates subject/object and union proposals with Region Proposal Network (RPN), which are sometimes derived from the ground-truth human annotations of the image. Each union proposal is made up of a subject, an object and a predicate ROI.} The predicate ROI is the box that tightly covers both the subject and the object. We can then obtain appearance, spatial information, label, depth, and mask for each object proposal using the feature representation, and for each predicate proposal we can obtain appearance, spatial, depth, and mask. These multimodal features are vectorized, combined, and refined in the third step of the Feature Refinement module using message passing mechanisms, attention mechanisms and visual translation embedding approaches. Finally, the classifiers are used to predict the categories of the predicates, and the scene graph is generated. 

In this section, SGG methods for 2D inputs will be reviewed and analyzed according to the following strategies.

\textcolor{black}{(1) Off-the-shelf object detectors can be used to detect subjects, objects and predicate ROIs. The first point to consider is how to utilize the multimodal features of the detected proposals. As a result, \textbf{Section 3.1.1} reviews and analyzes the use of \textbf{multimodal features}, including appearance, spatial, depth, mask, and label.}

\textcolor{black}{(2) A scene graph's compositionality is its most important characteristic, and can be seen as an elevation of its semantic expression from independent objects to visual phrases. A deeper meaning can, however, be derived from two aspects: the frequency of visual phrases and the common-sense constraints on relationship prediction. For example, when ``man", ``horse" and ``hat" are detected individually in an image, the most likely visual triplets are $\left\langle man, ride, horse \right\rangle$, $\left\langle man, wearing, hat \right\rangle$, etc. $\left\langle hat, on, horse \right\rangle$ is possible, though not common. But $\left\langle horse, wearing, hat \right\rangle$ is normally unreasonable. Thus, how to integrate \textbf{Prior Information} about visual phrases and \textbf{Commonsense Knowledge} will be the analyzed in \textbf{Section 3.1.2} and \textbf{Section 3.1.3}, respectively.}

\textcolor{black}{(3) A scene graph is a representation of visual relationships between objects, and it includes contextual information about those relationships. To achieve high-quality predictions, information must be fused between the individual objects or relationships. In a scene graph, message passing can be used to refine local features and integrate contextual information, while attention mechanisms can be used to allow the models to focus on the most important parts of the scene. Considering the large intra-class divergence and long-tailed distribution problems, visual translation embedding methods have been proposed to model relationships by interpreting them as translations operating on the
low-dimensional embeddings of the entities. Therefore, we categorize the related methods into \textbf{Message Passing}, \textbf{Attention Mechanism}, and \textbf{Visual Translation Embedding}, which will be deeply analyzed in \textbf{Section 3.1.4}, \textbf{Section 3.1.5} and \textbf{Section 3.1.6}, respectively.}

\subsubsection{Multimodal Features}
\textcolor{black}{The appearance features of the subject, object, and predicate ROIs make up the input of SGG methods, and affect SGG significantly. The rapid development of deep learning based object detection and classification has led to the use of many types of classical CNNs to extract appearance features from ROIs cropped from a whole image by bounding boxes or masks. Some CNNs even outperform humans when it comes to detecting/classifying objects based on appearance features. Nevertheless, only the appearance features of a subject, an object, and their union region are insufficient to accurately recognize the relationship of a subject-object pair. In addition to appearance features, semantic features of object categories or relations, spatial features of object candidates, and even contextual features, can also be crucial to understand a scene and can be used to improve the visual relationship detection performance. In this subsection, some integrated utilization methods of \textit{Appearance}, \textit{Semantic}, \textit{Spatial} and \textit{Context} features will be reviewed and analyzed.}

\textcolor{black}{\textbf{Appearance-Semantic Features:} A straightforward way to fuse semantic features is to concatenate the semantic word embeddings of object labels to the corresponding appearance features. As in \cite{lu2016visual}, there is another approach that utilizes language priors from semantic word embeddings to finetune the likelihood of a predicted relationship, dealing with the fact that objects and predicates independently occur frequently, even if relationship triplets are infrequent. Moreover, taking into account that the appearance of objects may profoundly change when they are involved in different visual relations, it is also possible to directly learn an appearance model to recognize richer-level visual composites, i.e., visual phrases\cite{sadeghi2011recognition}, as a whole, rather than detecting the basic atoms and then modeling their interactions.}

\textbf{Appearance-Semantic-Spatial Features:} \textcolor{black}{The spatial distribution of objects is not only a reflection of their position, but also a representation of their structural information.  A spatial distribution of objects is described by the properties of regions, which include positional relations, size relations, distance relations, and shape relations.} In this context, Zhu \emph{et al}.\cite{zhu2017visual} investigated how the spatial distribution of objects can aid in visual relation detection. Sharifzadeh \emph{et al}.\cite{sharifzadeh2019improving} used 3D information in visual relation detection by synthetically generating depth maps from an RGB-to-Depth model incorporated within relation detection frameworks. They extracted pairwise feature vectors for depth, spatial, label and appearance. 

\textcolor{black}{The subject and object come from different distributions. In response, Zhang \emph{et al}.\cite{zhang2017relationship} proposed a 3-branch Relationship Proposal Networks (Rel-PN) to produce a set of candidate boxes that represent subject, relationship, and object proposals. Then a proposal selection module selects the candidate pairs that satisfy the spatial constraints. The resulting pairs are fed into two separate network modules designed to evaluate the relationship compatibility using visual and spatial criteria, respectively. Finally, visual and spatial scores are combined with different weights to get the final score for predicates.} In another work\cite{zhang2018interpretable}, the authors added a semantic module to produce a semantic score for predicates, then all three scores are added up to obtain an overall score.
Liang \emph{et al}.\cite{liang2018visual} also considered three types of features and proposed to cascade the multi-cue based convolutional neural network with a structural ranking loss function. For an input image $x$, the feature representations of visual appearance cue, spatial location cue and semantic embedding cue are extracted for each relationship instance tuple. The learned features combined with multiple cues are further concatenated and fused into a joint feature vector through one fully connected layer. 

\textbf{Appearance-Semantic-Spatial-Context Features:} Previous studies typically extract features from a restricted object-object pair region and focus on local interaction modeling to infer the objects and pairwise relation. \textcolor{black}{For example, by fusing pairwise features, VIP-CNN \cite{li2017vip} captures contextual information directly. However, the global visual context beyond these pairwise regions is ignored, it may result in the loss of the chance to shrink the possible semantic space using the rich context. Xu \emph{et al}.\cite{xu2020scene} proposed a multi-scale context modeling method that can simultaneously discover and integrate the object-centric and region-centric contexts for inference of scene graphs in order to overcome the problem of large object/relation spaces. Yin \emph{et al}.\cite{yin2018zoom} proposed a Spatiality-Context-Appearance module to learn the spatiality-aware contextual feature representation.}

\textcolor{black}{\textbf{In summary}, appearance, semantics, spatial and contextual features all contribute to visual relationship detection from different perspectives. The integration of these multimodal features precisely corresponds to the human's multi-scale, multi-cue cognitive model. Using well-designed features, visual relationships will be detected more accurately so scene graphs can be constructed more accurately. }

\subsubsection{Prior Information}
\label{sec:Using_language_Priors}

The scene graph is a semantically structured description of a visual world. Intuitively, the SGG task can be regarded as a two-stage semantic tag retrieval process. Therefore, the determination of the relation category often depends on the labels of the participating subject and object. In Section 3.1, we discussed the \emph{compositionality} of a scene graph in detail. Although visual relationships are scene-specific, there are strong semantic dependencies between the relationship predicate \emph{r} and the object categories \emph{s} and \emph{o} in a relationship triplet (\emph{s}, \emph{p}, \emph{o}). 

\textcolor{black}{Data balance plays a key role in the performance of deep neural networks due to their data-dependent training process. However, because of the long-tailed distribution of relationships between objects, collecting enough training images for all relationships is time-consuming and too expensive \cite{jung2018visual, zhang2019large, dupty2020visual, yu2017visual}. Scene graphs should serve as an objective semantic representation of the state of a scene. We cannot arbitrarily assign the relationship of $\left\langle man, feeding, horse \right\rangle$ to the scene in Fig. 3(b) just because $\left\langle man, feeding, horse \right\rangle$ occurs more frequently than $\left\langle man, riding, horse \right\rangle$ in some datasets. However, in fact, weighting the probability output of relationship detection networks by statistical co-occurrences may improve the visual relationship detection performance on some datasets. We cannot deny the fact that human beings sometimes think about the world based on their experiences. As such, prior information, including \textbf{Statistical Priors} and \textbf{Language Priors}, can be regarded as a type of experience that allows neural networks to ``correctly understand" a scene more frequently. Prior information has already been widely used to improve performance of SGG networks.} 

\textbf{Statistical Priors:} \textcolor{black}{The simplest way to use prior knowledge is to think that an event should happen this time since it almost always does. This is called statistical prior. Baier \emph{et al}.\cite{baier2017improving} demonstrated how a visual statistical model could improve visual relationship detection. Their semantic model was trained using absolute frequencies that describe how often a triplet appears in the training data. Dai \emph{et al}.\cite{dai2017detecting} designed a deep relational network that exploited both spatial configuration and statistical dependency to resolve ambiguities during relationship recognition. Zellers \emph{et al}.\cite{zellers2018neural} analyzed the statistical co-occurrences between relationships and object pairs on the Visual Genome dataset and concluded that these statistical co-occurrences provided strong regularization for relationship prediction. } 

\textcolor{black}{Furthermore, Chen \emph{et al}.\cite{chen2019knowledge} formally represented this information and explicitly incorporated it into graph propagation networks to aid in scene graph generation.} For each object pair with predicted labels (a subject $o_i$ and an object $o_j$), they constructed a graph with a subject node, an object node, and $K$ relation nodes. Each node $v \in V = \{o_i, o_j, r_1, r_2, \ldots, r_K\}$ has a hidden state $h^t_v$ at timestep $t$. Let $m_{o_i o_j r_k}$ denote the statistical co-occurrence probability between $o_i$ and relation node $r_k$ as well as $o_j$ and relation node $r_k$. At timestep $t$, the relationship nodes aggregate messages from the object nodes, while object nodes aggregate messages from the relationship nodes:

\begin{equation}\label{eq:Routing_Network-messages}
\begin{aligned}
a_v^t = \begin{cases} 
\sum_{k = 1}^K m_{o_i o_j r_k} h_{r_k}^{t - 1}, & \text{if $v$ is an object node} \\
m_{o_i o_j r_k}(h_{o_i}^{t - 1} + h_{o_j}^{t - 1}), & \text{if $v$ is a relation node} \end{cases}
\end{aligned}
\end{equation}
Then, the hidden state $h^t_v$ is updated with $a_v^t$ and its previous hidden state by a gated mechanism.

\textcolor{black}{These methods, however, are data-dependent because their statistical co-occurrence probability is derived from training data. They do not contribute to the design of a universal SGG network. We believe that in the semantic space, language priors will be more useful.}

\begin{figure}[h] 
	\centering 
	\includegraphics[width=0.49\textwidth]{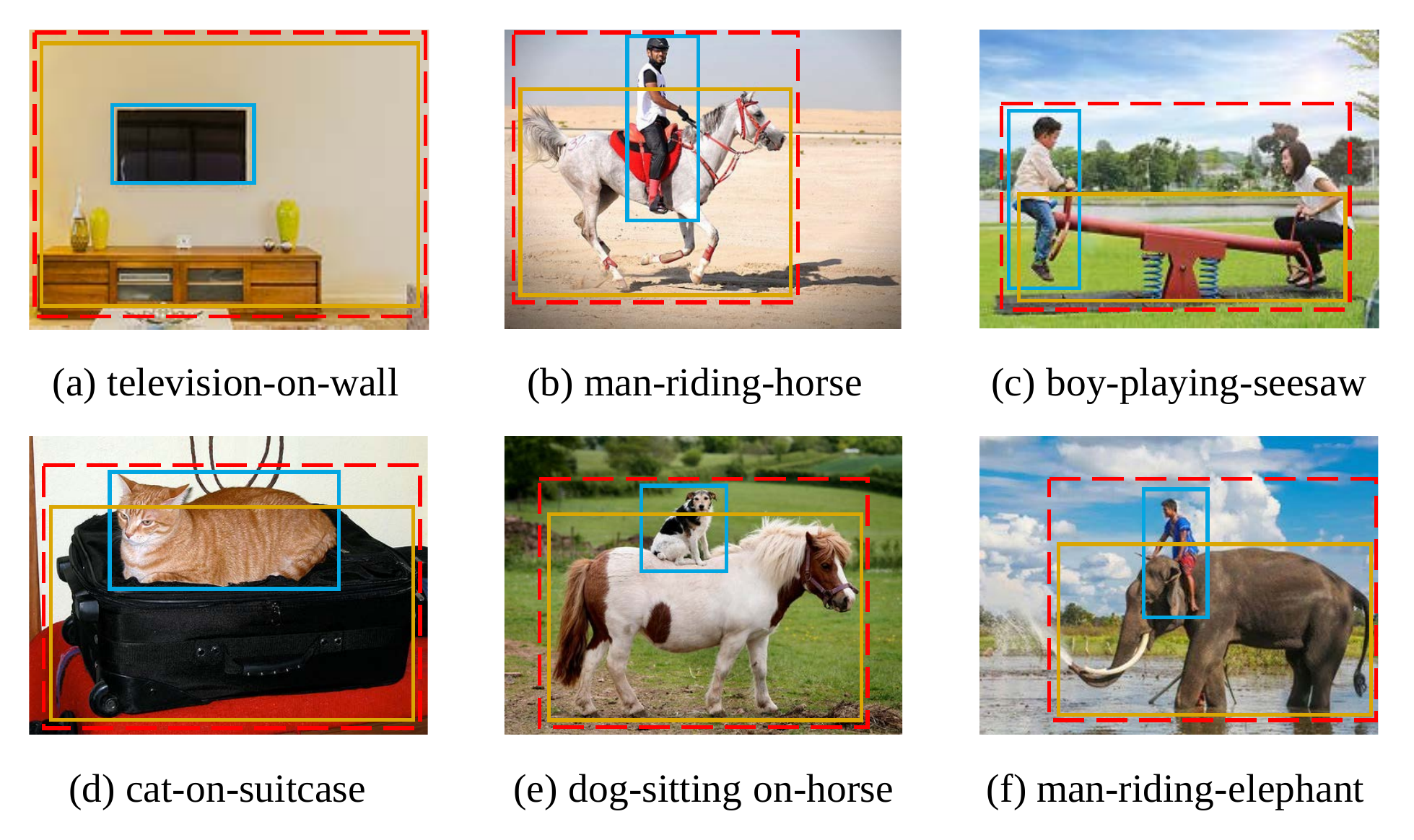} 
	\caption{\normalsize{Examples of the wide variety of visual relationships. The solid bounding boxes indicate the individual objects and the dash red bounding boxes denote a visual relationship.}}
	\label{fig:Various_Examples}  
\end{figure} 

\textbf{Language Priors:} \textcolor{black}{Human communication is primarily based on the use of words in a structured and conventional manner. Similarly, visual relationships are represented as triplets of words. Given the polysemy of words across different contexts, one cannot simply encode objects and predicates as indexes or bitmasks. The semantics of object and predicate categories should be used to deal with the polysemy in words. In particular, the following observations can be made.} \bfseries First\mdseries, the visual appearance of the relationships which has the same predicate but different agents varies greatly\cite{li2017vip}. For instance, the ``\emph{television-on-wall}" ({Fig.\ref{fig:Various_Examples}a}) and ``\emph{cat-on-suitcase}" ({Fig.\ref{fig:Various_Examples}d}) have the same predicate type ``\emph{on}", but they have distinct visual and spatial features. \bfseries Second\mdseries, the type of relations between two objects is not only determined by their relative spatial information but also through their categories. For example, the relative position between the kid and the horse ({Fig.\ref{fig:Various_Examples}b}) is very similar to the ones between the dog and the horse ({Fig.\ref{fig:Various_Examples}e}), but it is preferred to describe the relationship ``dog-sitting on-horse" rather than ``dog-riding-horse" in the natural language setting. It is also very rare to say ``person-sitting on-horse". On the other hand, the relationships between the observed objects are naturally based on our language knowledge. For example, we would like to use the expression ``sitting on" or ``playing" for seesaw but not ``riding" ({Fig.\ref{fig:Various_Examples}c}), even though it has a very similar pose as the one of the types ``riding" the horse in {Fig.\ref{fig:Various_Examples}b}. \bfseries Third\mdseries, relationships are semantically similar when they appear in similar contexts. That is, in a given context, i.e., an object pair, the probabilities of different predicates to describe this pair are related to their semantic similarity. For example, ``person-ride-horse" ({Fig.\ref{fig:Various_Examples}b}) is similar to ``person-ride-elephant" ({Fig.\ref{fig:Various_Examples}f}), since ``horse" and ``elephant" belong to the same animal category\cite{lu2016visual}. \textcolor{black}{It is therefore necessary to explore methods for utilizing language priors in the semantic space.}

Lu \emph{et al}.\cite{lu2016visual} proposed the first visual relationship detection pipeline, which leverages the language priors (LP) to finetune the prediction. They scored each pair of object proposals $\left\langle O_1, O_2\right\rangle$ using a visual appearance module and a language module. In the training phase, to optimize the projection function $f(.)$ such that it projects similar relationships closer to one another, they used a heuristic formulated as:

\begin{equation}\label{eq:LP_constant}
\begin{aligned}
constant = \frac{[f(r, W) - f(r\prime, W)]^2}{d(r, r\prime)}, \forall r, r\prime
\end{aligned}
\end{equation}
where $d(r, r\prime)$ is the sum of the cosine distances in word2vec space between the two objects and the predicates of the two relationships $r$ and $r\prime$. Similarly, Plesse \emph{et al}.\cite{plesse2018learning} computed the similarity between each neighbor $r\prime \in \{r_1, \ldots , r_K\}$ and the query $r$ with a softmax function:

\begin{equation}\label{eq:Prototypes_constant}
\begin{aligned}
constant = \frac{e^{-d(r,r\prime)^2}}{\sum_{j = 1}^K e^{-d(r,r_j)^2}}
\end{aligned}
\end{equation}
Based on this LP model, Jung \emph{et al}.\cite{jung2018visual} further summarized some major difficulties for visual relationship detection and performed a lot of experiments on all possible models with variant modules. 

\textcolor{black}{Liao \emph{et al}.\cite{liao2019natural} assumed that an inherent semantic relationship connects the two words in the triplet rather than a mathematical distance in the embedding space. They proposed to use a generic bi-directional RNN to predict the semantic connection between the participating objects in a relationship from the aspect of natural language. Zhang \emph{et al}.\cite{zhang2019large} used semantic associations to compensate for infrequent classes on a large and imbalanced benchmark with an extremely skewed class distribution. Their approach was to learn a visual and a semantic module that maps features from the two modalities into a shared space and then to employ the modified triplet loss to learn the joint visual and semantic embedding. As a result, Abdelkarim \emph{et al}.\cite{Abdelkarim2020LongtailVR} highlighted the long-tail recognition problem and adopted a weighted version of the softmax triplet loss above.}

From the perspective of collective learning on multi-relational data, Hwang \emph{et al}.\cite{Hwang2018TensorizeFA} designed an efficient multi-relational tensor factorization algorithm that yields highly informative priors. Analogously, Dupty \emph{et al}.\cite{dupty2020visual} learned conditional triplet joint distributions in the form of their normalized low rank non-negative tensor decompositions. 

In addition, some other papers have also tried to mine the value of language prior knowledge for relationship prediction. Donadello \emph{et al}.\cite{donadello2019compensating} encoded visual relationship detection with Logic Tensor Networks (LTNs), which exploit both the similarities with other seen relationships and background knowledge, expressed with logical constraints between subjects, relations and objects. In order to leverage the inherent structures of the predicate categories, Zhou \emph{et al}.\cite{zhou2020exploring} proposed to firstly build the language hierarchy and then utilize the Hierarchy Guided Feature Learning (HGFL) strategy to learn better region features of both the coarse-grained level and the fine-grained level. Liang \emph{et al}.\cite{liang2017deep} proposed a deep Variation-structured Reinforcement Learning (VRL) framework to sequentially discover object relationships and attributes in an image sample. Recently, Wen \emph{et al}.\cite{wen2020unbiased} proposed the Rich and Fair semantic extraction network (RiFa), which is able to extract richer semantics and preserve the fairness for relations with imbalanced distributions.

\textcolor{black}{\textbf{In summary}, statistical and language priors are effective in providing some regularizations, for visual relationship detection, derived from statistical and semantic spaces. However, additional knowledge outside of the scope of object and predicate categories, is not included. The human mind is capable of reasoning over visual elements of an image based on common sense. Thus, incorporating commonsense knowledge into SGG tasks will be valuable to explore.}

\subsubsection{Commonsense Knowledge}

\begin{figure}[ht] 
	\centering 
	\includegraphics[width=\linewidth]{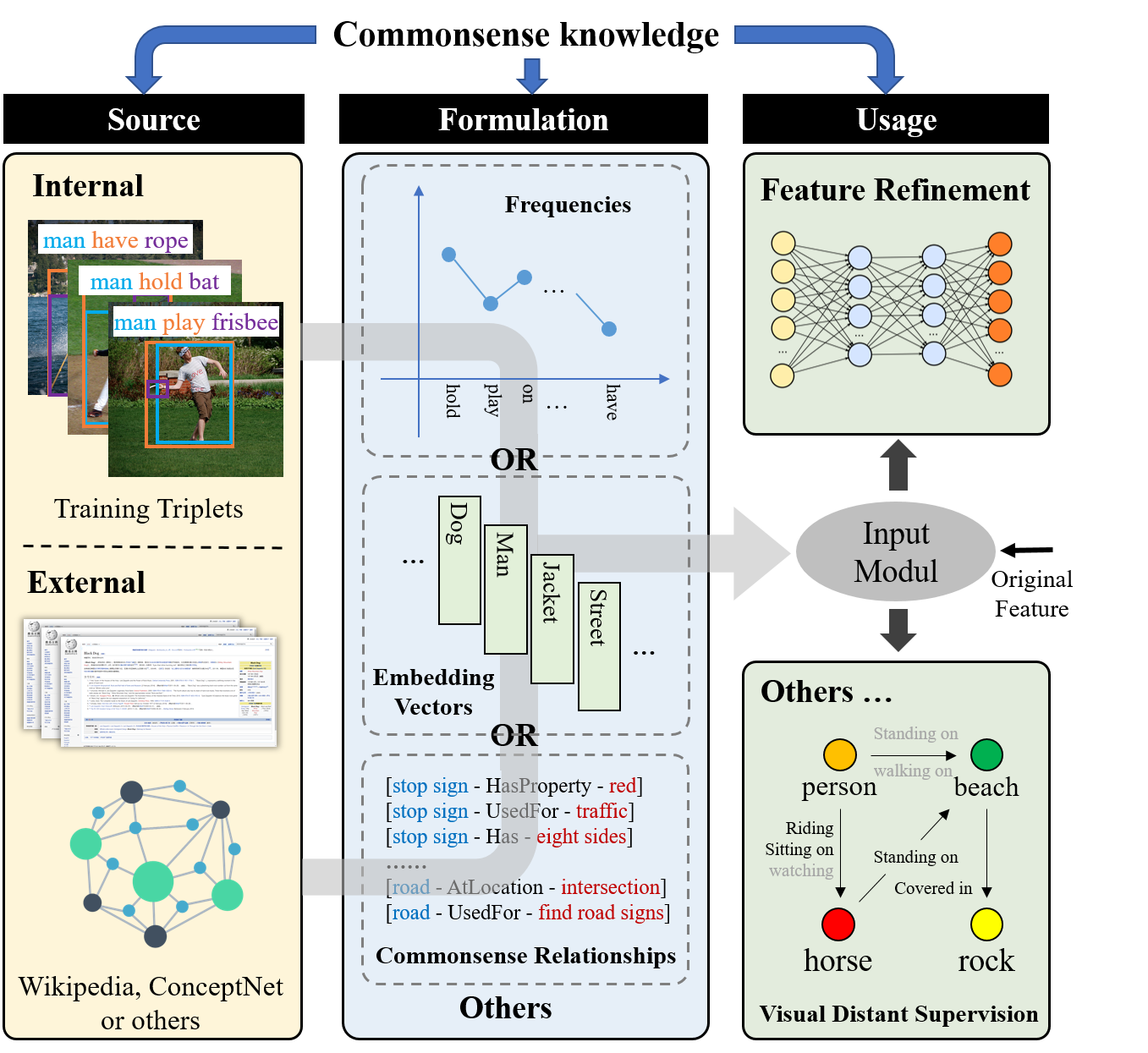} 
	\caption{\normalsize{Three basic sub-issues (source, formulation and usage) of commonsense knowledge applied to scene graph generation. }}
	\label{fig:Commonsense}  
\end{figure}

\textcolor{black}{As previously stated, there are a number of models which emphasize on the importance of language priors. However, due to the long tail distribution of relationships, it is costly to collect enough training data for all relationships\cite{yu2017visual}. We should therefore use knowledge beyond the training data to help generate scene graphs\cite{yang2017on}. Commonsense knowledge includes information about events that occur in time, about the effects of actions, about physical objects and how they are perceived, and about their properties and relationships with one another.} Researchers have proposed to extract commonsense knowledge to refine object and phrase features to improve generalizability of scene graph generation. In this section, we analyze three fundamental sub-issues of commonsense knowledge applied to SGG, i.e., the \textbf{Source}, \textbf{Formulation} and \textbf{Usage}, as illustrated in Fig. \ref{fig:Commonsense}. To be specific, the source of commonsense is generally divided into internal training samples\cite{chen2019soft,2021Multimodal}, external knowledge base\cite{gu2019scene} or both\cite{yu2017visual, zhan2019exploring}, and it can be transformed into different formulations\cite{2021Atom}. It is mainly applied in the feature refinement on the original feature or other typical procedures\cite{2021Visual}.

\textbf{Source:} Commonsense knowledge can be directly extracted from the local training samples. For example, Duan \emph{et al}.\cite{2021Multimodal} calculated the co-occurrence probability of object pairs, $p\left(o_{i} \mid o_{j}\right)$, and relationship in the presence of object pairs, $p\left(r_{k} \mid o_{i}, o_{j}\right)$, as the prior statistical knowledge obtained from the training samples of VG dataset, to assist reasoning and deal with the unbalanced distribution. However, considering the tremendous valuable information from the large-scale external bases, e.g., Wikipedia and ConceptNet, increasing efforts have been devoted to distill knowledge from these resources. 

Gu \emph{et al}.\cite{gu2019scene} proposed a knowledge-based module, which improves the feature refinement procedure by reasoning over a basket of commonsense knowledge retrieved from ConceptNet. Yu \emph{et al}.\cite{yu2017visual} introduced a Linguistic Knowledge Distillation Framework that obtains linguistic knowledge by mining from both training annotations (internal knowledge) and publicly available text, e.g., Wikipedia (external knowledge), and then construct a teacher network to distill the knowledge into a student network that predicts visual relationships from visual, semantic and spatial representations. Zhan \emph{et al}.\cite{zhan2019exploring} proposed a novel multi-modal feature based undetermined relationship learning network (MF-URLN), which extracts and fuses features of object pairs from three complementary modules: visual, spatial, and linguistic modules. The linguistic module provides two kinds of features: external linguistic features and internal linguistic features. The former is the semantic representations of subject and object generated by the pretrained word2vec model of Wikipedia 2014. The latter is the probability distributions of all relationship triplets in the training set according to the subject’s and object’s categories.

\textbf{Formulation:} Except from the actual sources of knowledge, it is also important to consider the formulation and how to incorporate the knowledge in a more efficient and comprehensive manner. As shown in several previous studies \cite{chen2019knowledge}, \cite{2021Multimodal}, the statistical correlation has been the most common formulation of knowledge. They employ the co-occurrence matrices on both the object pairs and the relationships in an explicit way. Similarly, Linguistic knowledge from \cite{zhan2019exploring} is modeled by a conditional probability that encodes the strong correlation between the object pair $\left\langle subj, obj\right\rangle$ and the predicate. However, Lin \emph{et al}.\cite{2021Atom} pointed out that they were generally composable, complex and image-specific which will lead to a poor learning improvement. Lin proposed Atom Correlation Based Graph Propagation (ACGP) for the scene graph generation task. The most significant one is to separate the relationship semantics to form new nodes and decompose the conventional multi-dependency reasoning path of $\left\langle subject, predicate, object\right\rangle$ into four different types of atom correlations, i.e., $\left\langle subject, object\right\rangle$, $\left\langle subject, predicate\right\rangle$, $\left\langle predicate, predicate\right\rangle$, $\left\langle predicate, object\right\rangle$, which are much more flexible and easier to learn. It consequently results in four kinds of knowledge graph, then the information propagation is performed using the graph convolutional network (GCN) with the guidance of the knowledge graphs, to produce the evolved node features. 

\textbf{Usage:} In general, the commonsense knowledge is used as guidance on the original feature refinement for most cases \cite{2021Multimodal}, \cite{gu2019scene}, \cite{2021Atom}, but there are a lot of attempts to implement it and it contributes from a different aspect on the scene graph model. Yao \emph{et al}.\cite{2021Visual} demonstrated a framework which can train the scene graph models in an unsupervised manner, based on the knowledge bases extracted from the triplets of Web-scale image captions. The relationships from the knowledge base are regarded as the potential relation candidates of corresponding pairs. The first step is to align the knowledge base with images and initialize the probability distribution $D_{S}$ for each candidate: 

\begin{equation}\label{eq:VD-pdistrbution}
	\begin{aligned}
		d=\Psi(s, o, \Lambda)
	\end{aligned}
\end{equation}
where the $(s,o)$ denotes the corresponding object pair proposed by the detector, $\Lambda$ represents the knowledge base and $\Psi(\cdot)$ is the alignment procedure. \textbf{$d$} is a vector, where $d_{i}=1$ if the relation $r_{i}$ belongs to the set of retrieved relation labels from knowledge base, otherwise 0. In every non-initial iteration $t(t>1)$, this distribution will be constantly updated by the convex combination of the internal prediction from the scene graph model and the external semantic signals. 

Inspired by a hierarchical reasoning from the human's prefrontal cortex, Yu \emph{et al}.\cite{2021CogTree} built a Cognition Tree (CogTree) for all the relationship categories in a coarse-to-fine manner. For all the samples with the same ground-truth relationship class, they predicted the relationships by a biased model and calculated the distribution of the predicted label frequency, based on what the hierarchical structure of CogTree can be consequently built. The tree can be divided into four layers (root, concept, coarse-fine, fine-grained) and progressively divides the coarse concepts with a clear distinction into fined-grained relationships which share similar features. On the basis of this structure, a model-independent loss function, tree-based class-balanced (TCB) loss is introduced in the training procedure. This loss can suppress the inter-concept and intra-concept noises in a hierarchical way, which finally contribute to an unbiased scene graph prediction.

Recently, Zareian \emph{et al}.\cite{zareian2020bridging} proposed a Graph Bridging Network (GB-NET). This model is based on an assumption that a scene graph can be seen as an image-conditioned instantiation of a commonsense knowledge graph. They generalized the formulation of scene graphs into knowledge graphs where predicates are nodes rather than edges and reformulated SGG from object and relation classification into graph linking. The GB-NET is an iterative process of message passing inside a heterogeneous graph. It consists of a commonsense graph and an initialized scene graph connected by some \emph{bridge} edges. The commonsense graph is made up of commonsense entity nodes (CE) and commonsense predicate nodes (CP), and the commonsense edges are compiled from three sources, WordNet, ConceptNet, and the Visual Genome training set. The scene graph is initialized with scene entity nodes (SE), \emph{i.e.}, detected objects, and scene predicate nodes (SP) for all entity pairs. Its goal is to create \emph{bridge} edges between the two graphs that connect each instance (SE and SP node) to its corresponding class (CE and CP node).

Another work by Zareian \emph{et al}.\cite{zareian2020learning} perfectly match the discussion of this section. It points out two specific issues of current researches on knowledge-based SGG methods: \textbf{(1)} external source of commonsense tend to be incomplete and inaccurate; \textbf{(2)} statistics information such as co-occurrence frequency is limited to reveal the complex, structured patterns of commonsense. Therefore, they proposed a novel mathematical formalization on visual commonsense and extracted it based on a global-local attention multi-heads transformer. It is implemented by training the encoders on the corpus of annotated scene graphs to predict the missing elements of a scene. Moreover, to compensate the disagreement between commonsense reasoning and visual prediction, it disentangles the commonsense and perception into two separate trained models and builds a cascaded fusion architecture to balance the results. The commonsense is then used to adjust the final prediction.

\textcolor{black}{As the main characteristics of commonsense knowledge, external large-scale knowledge bases and specially-designed formulations of statistical correlations have drawn considerable attention in recent years. However, \cite{2021Visual}, \cite{2021CogTree} have demonstrated that, except from feature refinement, commonsense knowledge can also be useful in different ways. Due to its graph-based structure and enriched information, commonsense knowledge may boost the reasoning process directly. Its graph-based structure makes it very important to guide the message passing on GNN- and GCN-based scene graph generation methods.}

\subsubsection{Message Passing}

\textcolor{black}{A scene graph consists not only of individual objects and their relations, but also of contextual information surrounding and forming those visual relationships. From an intuitive perspective, individual predictions of objects and relationships are influenced by their surrounding context. Context can be understood on three levels.  \bfseries First\mdseries, for a triplet, the predictions of different phrase components depend on each other. This is the compositionality of a scene graph. \bfseries Second\mdseries, the triplets are not isolated. Objects which have relationships are semantically dependent, and relationships which partially share object(s) are also semantically related to one another. \bfseries Third\mdseries, visual relationships are scene-specific, so learning feature representations from a global view is helpful when predicting relationships. Therefore, message passing between individual objects or triplets are valuable for visual relationship detection.}

\textcolor{black}{Constructing a high-quality scene graph relies on a prior layout structure of proposals (objects and unions). There are four forms of layout structures: triplet set, chain, tree and fully-connected graph. Accordingly, RNN and its variants (LSTM, GRU) as sequential models are used to encode context for chains while TreeLSTM\cite{tai2015improved} for trees and GNN (or CRF)\cite{kipf2016semi, hamilton2017inductive, velivckovic2017graph} for fully-connected graphs.}

\begin{figure}[ht] 
	\centering 
	\includegraphics[width=\linewidth]{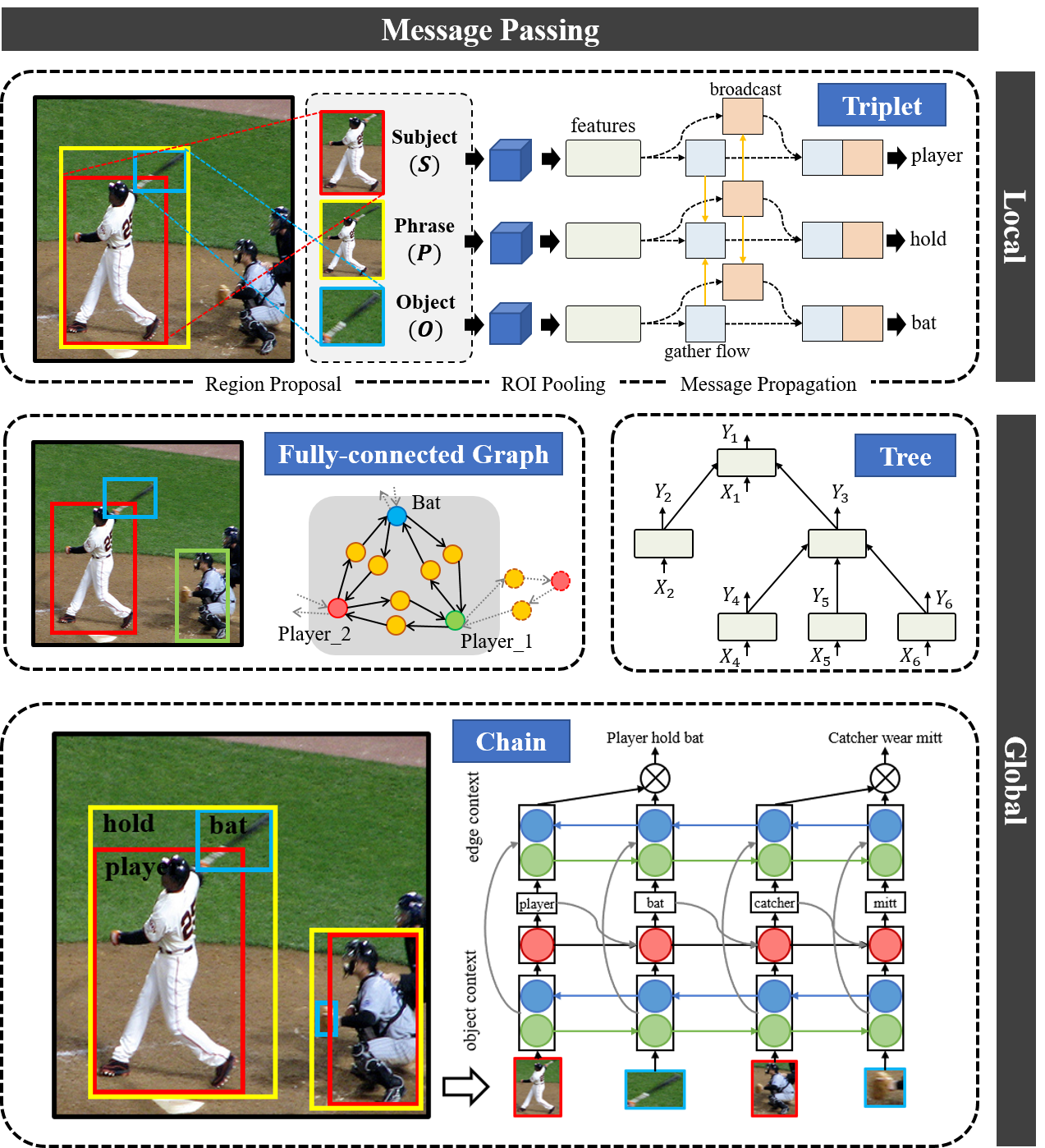} 
	\caption{\normalsize{Examples of two different types of message passing methods, i.e., local propagation within triplet items\cite{li2017vip} and global propagation across all the elements. The global items, according to layout structures, can be further divided into the following forms: fully-connected graph\cite{kipf2016semi, hamilton2017inductive}, chain\cite{shin2018deep, chen2019panet} and tree\cite{tang2019learning, tai2015improved}.}}
	\label{fig:Messagepassing}  
\end{figure}

\textcolor{black}{Basically, features and messages are passed between elements of a scene graph, including objects and relationships. To refine object features and extract phrase features, several models rely on a variety of message passing techniques. Our discussion in the subsections below is structured around two key perspectives: \textbf{local} propagation within triplet items and \textbf{global} propagation across all the elements, as illustrated in Fig.\ref{fig:Messagepassing}.}

\textcolor{black}{\textbf{Local Message Passing Within Triplets:} Generally, features of the subject, predicate and object proposals are extracted for each triplet, and the information fusion within the triplets contribute to refine features and recognize visual relationships. }ViP-CNN, proposed by Li \emph{et al}.\cite{li2017vip}, is a phrase-guided visual relationship detection framework, which can be divided into two parts: \emph{triplet proposal} and \emph{phrase recognition}. In phrase detection, for each triplet proposal, there are three feature extraction branches for subject, predicate and object, respectively. The phrase-guided message passing structure (PMPS) is introduced to exchange the information between branches. Dai \emph{et al}.\cite{dai2017detecting} proposed an effective framework called Deep Relational Network (DR-Net), which uses Faster RCNN to locate a set of candidate objects. Through multiple inference units, who capture the statistical relations between triplet components, the DR-Net outputs the posterior probabilities of \emph{s}, \emph{r}, and \emph{o}. At each step of the iterative updating procedure, it takes in a fixed set of inputs, \emph{i.e.} the observed features $x_s$, $x_r$, and $x_o$, and refines the estimates of posterior probabilities. Another interesting model is Zoom-Net\cite{yin2018zoom}, which propagates spatiality-aware object features to interact with the predicate features and broadcasts predicate features to reinforce the features of subject and object. The core of Zoom-Net is a Spatiality-Context-Appearance Module, which consists of two spatiality-aware feature alignment cells for message passing between the different components of a triplet. 

\textcolor{black}{The local message passing within triplets ignores the surrounding context, while the joint reasoning with contextual information can often resolve ambiguities caused by local predictions made in isolation. The passing of global messages across all elements enhances the ability to detect finer visual relationships.}

\textcolor{black}{\textbf{Global Message Passing Across All Elements:} Considering that objects that have visual relationships are semantically related to each other, and that relationships which partially share objects are also semantically related, passing messages between related elements can be beneficial. Learning feature representation from a global view is helpful to scene-specific visual relationship detection. Scene graphs have a particular structure, so message passing on the graph or subgraph structures is a natural choice. Chain-based models (such as RNN or LSTM) can also be used to encode contextual cues due to their ability to represent sequence features. When taking into consideration the inherent parallel/hierarchical relationships between objects, dynamic tree structures can also be used to capture task-specific visual contexts. In the following subsections, message passing methods will be analyzed according to the three categories described below.}

\textcolor{black}{\textbf{\textit{Message Passing on Graph Structures. }}Li \emph{et al}.\cite{li2017scene} developed an end-to-end Multi-level Scene Description Network (MSDN), in which message passing is guided by the dynamic graph constructed from objects and caption region proposals. In the case of a phrase proposal, the message comes from a caption region proposal that may cover multiple object pairs, and may contain contextual information with a larger scope than a triplet.} For comparison, the Context-based Captioning and Scene Graph Generation Network (C2SGNet)\cite{shin2018deep} also simultaneously generates region captions and scene graphs from input images, but the message passing between phrase and region proposals is unidirectional, i.e., the region proposals requires additional context information for the relationships between object pairs. Moreover, in an extension of MSDN model, Li \emph{et al}.\cite{li2018factorizable} proposed a subgraph-based scene graph generation approach called Factorizable Network (F-Net), where the object pairs referring to the similar interacting regions are clustered into a subgraph and share the phrase representation. F-Net clusters the fully-connected graph into several subgraphs to obtain a factorized connection graph by treating each subgraph as a node, and passing messages between subgraph and object features along the factorized connection graph with a Spatial-weighted Message Passing (SMP) structure for feature refinement.

\textcolor{black}{Even though MSDN and F-Net extended the scope of message passing, a subgraph is considered as a whole when sending and receiving messages.} Liao \emph{et al}.\cite{liao2019exploring} proposed semantics guided graph relation neural network (SGRNN), in which the target and source must be an object or a predicate within a subgraph. It first establishes an undirected fully-connected graph by associating any two objects as a possible relationship. Then, they remove the connections that are semantically weakly dependent, through a semantics guided relation proposal network (SRePN), and a semantically connected graph is formed. To refine the feature of a target entity (object or relationship), source-target-aware message passing is performed by exploiting contextual information from the objects and relationships that the target is semantically correlated with for feature refinement. The scope of messaging is the same as Feature Inter-refinement of objects and relations in \cite{gu2019scene}.

\textcolor{black}{Several other techniques consider SGG as a graph inference process because of its particular structure. By considering all other objects as carriers of global contextual information for each object, they will pass messages to each other’s via a fully-connected graph.} However, inference on a densely connected graph is very expensive. As shown in previous works\cite{krahenbuhl2011efficient, long2015fully}, dense graph inference can be approximated by \bfseries mean field \mdseries in Conditional Random Fields (CRF). Moreover, Johnson \emph{et al}.\cite{Johnson2015Image} designed a CRF model that reasons about the connections between an image and its ground-truth scene graph, and use these scene graphs as queries to retrieve images with similar semantic meanings. Zheng \emph{et al}. \cite{liang2016semantic, zheng2015conditional} combines the strengths of CNNs with CRFs, and formulates mean-field inference as Recurrent Neural Networks (RNN). Therefore, it is reasonable to use CRF or RNN to formulate a scene graph generation problem\cite{dai2017detecting, cong2018scene}. 

\textcolor{black}{Further, there are some other relevant works which proposed modeling methods based on a pre-determined graph.} Hu \emph{et al}.\cite{Hu2019NeuralMP} explicitly model objects and interactions by an interaction graph, a directed graph built on object proposals based on the spatial relationships between objects, and then propose a message-passing algorithm to propagate the contextual information. Zhou \emph{et al}.\cite{zhou2019visual} mined and measured the relevance of predicates using relative location and constructed a location-based Gated Graph Neural Network (GGNN) to improve the relationship representation. Chen \emph{et al}.\cite{chen2019knowledge} built a graph to associate the regions and employed a graph neural network to propagate messages through the graph. Dornadula \emph{et al}.\cite{dornadula2019visual} initialized a fully connected graph, \emph{i.e.,} all objects are connected to all other objects by all predicate edges, and updated their representation using message passing protocols within a well-designed graph convolution framework. Zareian \emph{et al}.\cite{zareian2020bridging} formed a heterogeneous graph by using some bridge edges to connect a commonsense graph and initialized a fully connected graph. They then employed a variant of GGNN to propagate information among nodes and updated node representations and bridge edges. Wang \emph{et al}.\cite{wang2019exploring} constructed a virtual graph with two types of nodes (objects $v_i^o$ and relations $v_{ij}^r$) and three types of edges ($\left\langle v_i^o, v_j^o\right\rangle$, $\left\langle v_i^o, v_{ij}^r\right\rangle$ and $\left\langle v_{ij}^r, v_j^o,\right\rangle$), and then refined representations for objects and relationships with an explicit message passing mechanism.

\textcolor{black}{\textit{\textbf{Message Passing on Chain Structures.}} Dense graph inference can be approximated by mean fields in CRF, and it can also be dealt with using an RNN-based model. }Xu \emph{et al}.\cite{xu2017scene} generated structured scene representation from an image, and solved the graph inference problem using GRUs to iteratively improve its predictions via message passing. This work is considered as a milestone in scene graph generation, demonstrating that RNN-based models can be used to encode the contextual cues for visual relationship recognition. At this point, Zellers \emph{et al}.\cite{zellers2018neural} presented a novel model, Stacked Motif Network (MOTIFNET), which uses LSTMs to create a contextualized representation of each object. Dhingra \emph{et al}.\cite{dhingra2021bgt} proposed an object communication module based on a bi-directional GRU layer and used two different transformer encoders to further refine the object features and gather information for the edges. The Counterfactual critic Multi-Agent Training (CMAT) approach\cite{chen2019counterfactual} is another important extension where an agent represents a detected object. Each agent communicates with the others for $T$ rounds to encode the visual context. In each round of communication, an LSTM is used to encode the agent interaction history and extracts the internal state of each agent. 

\textcolor{black}{Many other message passing methods based on RNN have been developed.} Chen \emph{et al}.\cite{chen2019panet} used an RNN module to capture instance-level context, including objects co-occurrence, spatial location dependency and label relation. Dai \emph{et al}.\cite{dai2019visual} used a Bi-directional RNN and Shin \emph{et al}.\cite{shin2018deep} used Bi-directional LSTM as a replacement. Masui \emph{et al}.\cite{masui2018recurrent} proposed three triplet units (TUs) for selecting a correct SPO triplet at each step of LSTM, while achieving class–number scalability by outputting a single fact without calculating a score for every combination of SPO triplets. 

\textcolor{black}{\textit{\textbf{Message Passing on Tree Structures.}} As previously stated,  graph and chain structures are widely used for message passing. However, these two structures are sub-optimal. Chains are oversimplified and may only capture simple spatial information or co-occurring information. Even though fully-connected graphs are complete, they do not distinguish between hierarchical relations.} Tang \emph{et al}.\cite{tang2019learning} constructed a dynamic tree structure, dubbed VCTREE, that places objects into a visual context, and then adopted bidirectional TreeLSTM to encode  the visual contexts. \textcolor{black}{VCTREE has several advantages over chains and graphs, such as hierarchy, dynamicity, and efficiency.} VCTREE construction can be divided into three stages: \bfseries (1) \mdseries learn a score matrix $S$, where each element is defined as the product of the object correlation and the pairwise task-dependency; \bfseries (2) \mdseries obtain a maximum spanning tree using the Prim's algorithm, with a root $i$ satisfying $arg max_i \sum_{j\neq i} S_{ij}$;  \bfseries (3) \mdseries convert the multi-branch tree into an equivalent binary tree (\emph{i.e.,} VCTREE) by changing non-leftmost edges into right branches. The ways of context encoding and decoding for objects and predicates are similar to \cite{zellers2018neural}, but they replace LSTM with TreeLSTM. In \cite{zellers2018neural}, Zellers \emph{et al}. tried several ways to order the bounding regions in their analysis. Here, we can see the tree structure in VCTREE as another way to order the bounding regions. 

\subsubsection{Attention Mechanisms}

\begin{figure}[ht] 
	\centering 
	\includegraphics[width=\linewidth]{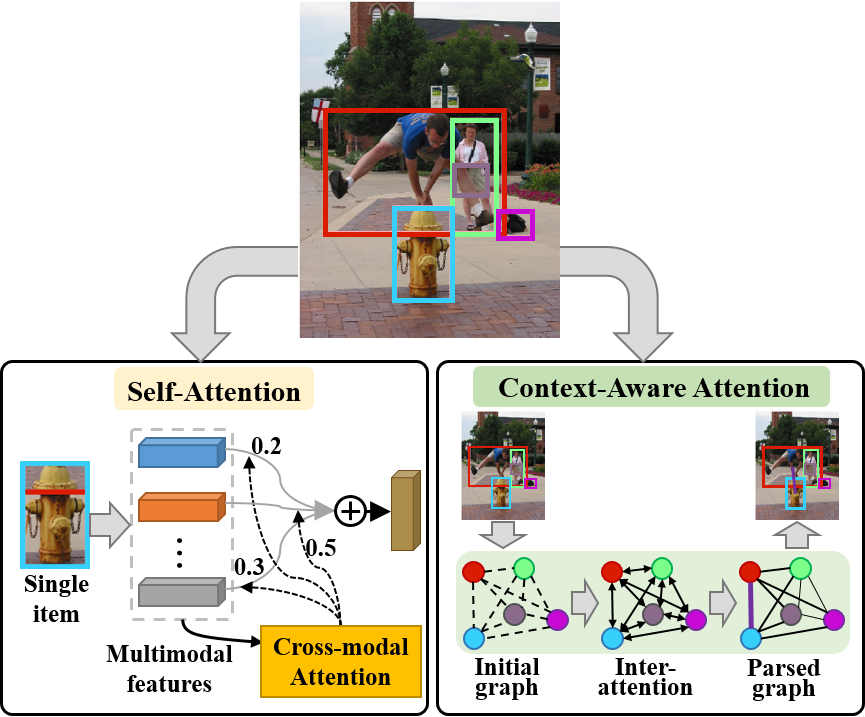} 
	\caption{\normalsize{Two kinds of attention mechanisms in SGG: (1) Self-Attention mechanism\cite{zheng2019visual} aggregates multimodal features of an object to generate a more comprehensive representation. (2) Context-Aware Attention\cite{yang2018graph} learns the contextual features using graph parsing.}}
	\label{fig:Attention}  
\end{figure}

\textcolor{black}{Attention mechanisms flourished soon after the success of Recurrent Attention Model (RAM)\cite{mnih2014recurrent} for image classification. They enable models to focus on the most significant parts of the input \cite{bahdanau2014neural}. With scene graph generation, as with iterative message passing models, there are two objectives: refine local features and fuse contextual information. On the basic framework shown in Fig.\ref{fig:scene-graph-generation-process}, attention mechanisms can be used both at the stage of feature representation and at the stage of feature refinement. At the feature representation stage, attention can be used in the spatial domain, channel domain or their mixed domain to produce a more precise appearance representation of object regions and unions of object-pairs. At the feature refinement stage, attention is used to update each object and relationship representation by integrating contextual information. Therefore, this section will analyze two types of attention mechanisms for SGG (as illustrated in Fig. \ref{fig:Attention}), namely, \textbf{Self-Attention} and \textbf{Context-Aware Attention} mechanisms.}

\textbf{Self-Attention Mechanisms.} \textcolor{black}{Self-Attention mechanism aggregates multimodal features of an object to generate a more comprehensive representation.} Zheng \emph{et al}.\cite{zheng2019visual} proposed a multi-level attention visual relation detection model (MLA-VRD), which uses multi-stage attention for appearance feature extraction and multi-cue attention for feature fusion. In order to capture discriminative information from the visual appearance, the channel-wise attention is applied in each convolutional block of the backbone network to improve the representative ability of low-level appearance features, and the spatial attention learns to capture the salient interaction regions in the union bounding box of the object pair. The multi-cue attention is designed to combine appearance, spatial and semantic cues dynamically according to their significance for relation detection.

In another work, Zhou \emph{et al}.\cite{zhou2019visual-attention} combined multi-stage and multi-cue attention to structure the Language and Position Guided Attention module (LPGA), where language and position information are exploited to guide the generation of more efficient attention maps. Zhuang \emph{et al}.\cite{zhuang2017towards} proposed a context-aware model, which applies an attention-pooling layer to the activations of the \emph{conv5\_3} layer of VGG-16 as an appearance feature representation of the union region. For each relation class, there is a corresponding attention model imposed on the feature map to generate a relation class-specific attention-pooling vectors. Han \emph{et al}.\cite{han2018visual} argued that the context-aware model pays less attention to small-scale objects. Therefore, they proposed the Vision Spatial Attention Network (VSA-Net), which employs a two-dimensional normal distribution attention scheme to effectively model small objects. The attention is added to the corresponding position of the image according to the spatial information of the Faster R-CNN outputs. Kolesnikov \emph{et al}.\cite{kolesnikov2019detecting} proposed the Box Attention and incorporated box attention maps in the convolutional layers of the base detection model. 

\textbf{Context-Aware Attention Mechanisms.} \textcolor{black}{Context-Aware Attention learns the contextual features using graph parsing.} Yang \emph{et al}.\cite{yang2018graph} proposed Graph R-CNN based on graph convolutional neural network (GCN)\cite{kipf2016semi}, which can be factorized into three logical stages: \bfseries(1) \mdseries produce a set of localized object regions, \bfseries(2) \mdseries utilize a relation proposal network (RePN) to learns to efficiently compute relatedness scores between object pairs, which are used to intelligently prune unlikely scene graph connections, and \bfseries(3) \mdseries apply an attentional graph convolution network (aGCN) to propagate a higher-order context throughout the sparse graph. In the aGCN, for a target node $i$ in the graph, the representations of its neighboring nodes $\{z_j | j \in \mathcal{N}(i)\}$ are first transformed via a learned linear transformation $W$. Then, these transformed representations are gathered with predetermined weights $\alpha$, followed by a nonlinear function $\sigma$ (ReLU). This layer-wise propagation can be written as:

\begin{equation}\label{eq:Graph R-CNN}
\begin{aligned}
z_i^{(l+1)} = \sigma\Big(z_i^{(l)} + \sum_{j \in \mathcal{N}(i)} \alpha_{ij} W z_j^{(l)}\Big)
\end{aligned}
\end{equation}
The attention $\alpha_{ij}$ for node $i$ is:

\begin{equation}\label{eq:Graph R-CNN_attention}
\begin{aligned}
\alpha_{ij} = \text{softmax}(w_h^T\sigma(W_a[z_i^{(l)}, z_j^{(l)}]))
\end{aligned}
\end{equation}
where $w_h$ and $W_a$ are learned parameters and $[\cdot ; \cdot]$ is the concatenation operation. From the derivation, it can be seen that the aGCN is similar to Graph Attention Network (GAT)\cite{velivckovic2017graph}. In conventional GCN, the connections in the graph are known and the coefficient vectors $\alpha_{ij}$ are preset based on the symmetrically normalized adjacency matrix of features. 

Qi \emph{et al}.\cite{qi2019attentive} also leveraged a graph self-attention module to embed entities, but the strategies to determine connection (i.e., edges that represent relevant object pairs are likely to have relationships) are different from the RePN, which uses the multi-layer perceptron (MLP) to learn to efficiently estimate the \emph{relatedness} of an object pair, where the adjacency matrix is determined with the space position of nodes. Lin \emph{et al}.\cite{Lin2020GPSNetGP} designed a
direction-aware message passing (DMP) module based on GAT to enhances the node feature with node-specific contextual information. Moreover, Zhang \emph{et al}.\cite{zhang2019relationship} used context-aware attention mechanism directly on the fully-connected graph to refine object region feature and performed comparative experiments of Soft-Attention and Hard-Attention in ablation studies. Dornadula \emph{et al}.\cite{dornadula2019visual} introduced another interesting GCN-based attention model, which treats predicates as learned semantic and spatial functions that are trained within a graph convolution network on a fully connected graph where object representations form the nodes and the predicate functions act as edges.

\subsubsection{Visual Translation Embedding}

Each visual relation involves subject, object and predicate, resulting in a greater skew of rare relations, especially when the co-occurrence of some pairs of objects is infrequent in the dataset. Some types of relations contain very limited examples. The \bfseries long-tail \mdseries problem heavily affects the scalability and generalization ability of learned models. Another problem is the \bfseries large intra-class divergence \mdseries\cite{zhou2018object}, i.e., relations that have the same predicate but from which different subjects or objects are essentially different. Therefore, there are two challenges for visual relationship detection models. \bfseries First\mdseries, is the right representation of visual relations to \textbf{handle the large variety in their appearance}, which depends on the involved entities. \bfseries Second\mdseries, is to  handle the scarcity of training data for zero-shot visual relation triplets. Visual embedding approaches aim at learning a compositional representation for subject, object and predicate by learning separate visual-language embedding spaces, where each of these entities is mapped close to the language embedding of its associated annotation. By constructing a mathematical relationship of visual-semantic embeddings for subject, predicate and object, an end-to-end architecture can be built and trained to learn a visual translation vector for prediction. In this section, we divide the visual translation embedding methods according to the translations (as illustrated in Fig.7), including \textbf{Translation between Subject and Object}, and \textbf{Translation among Subject, Object and Predicate}.

\begin{figure}[ht] 
	\centering 
	\includegraphics[width=\linewidth]{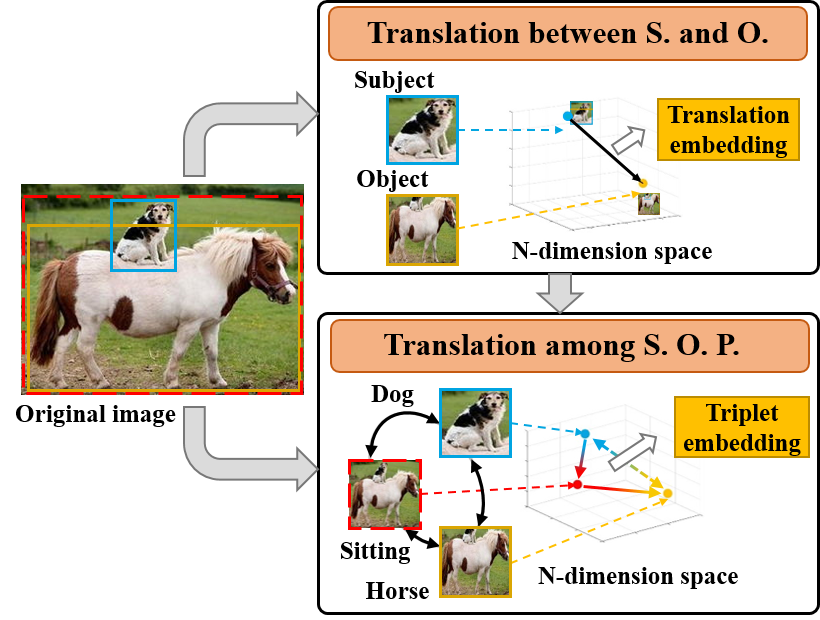} 
	\caption{\normalsize{Two types of visual translation embedding approaches according to whether to embed the predicate into N-dimensional space \cite{gkanatsios2019attention} or not \cite{zhang2017visual},  besides the subject and object embedding.}}
	\label{fig:TransE}  
\end{figure}

\textbf{Translation Embedding between Subject and Object:} Translation-based models in knowledge graphs are good at learning embeddings, while also preserving the structural information of the graph\cite{bordes2013translating, wang2014knowledge, ji2016knowledge}. Inspired by Translation Embedding (TransE)\cite{bordes2013translating} to represent large-scale knowledge bases, Zhang \emph{et al}.\cite{zhang2017visual} proposed a Visual Translation Embedding network (VTransE) which places objects in a low-dimensional relation space, where a relationship can be modeled as a simple vector translation, i.e., $\text{subject} + \text{predicate} \approx \text{object}$. Suppose $ x_s, x_o \in \mathbbm{R}^M $ are the \emph{M}-dimensional features, VTransE learns a relation translation vector $\bm{t}_p \in \mathbbm{R}^r$ ($r \ll M$) and two projection matrices $\bm{W}_s, \bm{W}_o \in \mathbbm{R}_{r \times M}$ from the feature space to the relation space. The visual relation can be represented as:

\begin{equation}\label{eq:VTransE_Net}
\begin{aligned}
\bm{W}_s x_s + t_p \approx \bm{W}_o x_o
\end{aligned}
\end{equation}
The overall feature $x_s$ or $x_o$ is a weighted concatenation of three features: semantic, spatial and appearance. The semantic information is an $(N + 1)$-d vector of object classification probabilities (i.e., $N$ classes and $1$ background) from the object detection network, rather than the word2vec embedding of label.

\textbf{Translation Embedding between Subject, Object and Predicate:} \textcolor{black}{In an extension of VTransE, Hung \emph{et al}.\cite{hung2020contextual} proposed the Union Visual Translation Embedding network (UVTransE), which learns three projection matrices $\bm{W}_s$, $\bm{W}_o$, $\bm{W}_u$ which map the respective feature vectors of the bounding boxes enclosing the \emph{subject}, \emph{object}, and \emph{union} of subject and object into a common embedding space, as well as translation vectors $\bm{t}_p$(to be consistent with VTransE) in the same space corresponding to each of the predicate labels that are present in the dataset.} Another extension is ATR-Net (Attention-Translation-Relation Network), proposed by Gkanatsios \emph{et al}.\cite{gkanatsios2019attention} which projects the visual features from the subject, the object region and their union into a score space as $S$, $O$ and $P$ with multi-head language and spatial attention guided. Let $A$ denotes the attention matrix of all predicates, Eq. \ref{eq:VTransE_Net} can be reformulated as:

\begin{equation}\label{eq:ART_Net}
\begin{aligned}
\bm{W}_p(A) x_p \approx \bm{W}_o(A) x_o - \bm{W}_s(A) x_s 
\end{aligned}
\end{equation}
Contrary to VTransE, the authors do not directly align $P$ and $O - S$ by minimizing $||P + S - O||$, instead, they create two separate score spaces for predicate classification ($p$) and object relevance ($r$), respectively and impose loss constraints, $\mathcal{L}^e_P$ and $\mathcal{L}^e_{OS}$ ($e$ \text{can be} $p$ \text{or} $r$), to force both $P$ and $O - S$ to match with the task’s ground-truth as follows:

\begin{equation}\label{eq:ART_Net_task_loss}
\begin{aligned}
\mathcal{L}^e(W^e) = \mathcal{L}^e_f(W^e) + \mathcal{L}^e_P(W^e_P) + \mathcal{L}^e_{OS}(W^e_O, W^e_S)
\end{aligned}
\end{equation}
Subsequently, Qi \emph{et al}.\cite{qi2019attentive} introduced a semantic transformation module into their network structure to represent $\left\langle \emph{S}, \emph{P}, \emph{O} \right\rangle$ in the semantic domain. This module leverages both the visual features (\emph{i.e.,} $f_i$, $f_j$ and $f_{ij}$) and the semantic features (\emph{i.e.,} $v_i$, $v_j$ and $v_{ij}$) that are concatenated and projected into a common semantic space to learn the relationship between pair-wise entities. L2 loss is used to guide the learning process:

\begin{equation}\label{eq:ARN_STM_Loss}
	\begin{aligned}
		\mathcal{L} = ||W_3 \cdot [f_j, v_o] - (W_1 \cdot [f_i, v_i] + W_2 \cdot [f_{ij}, v_{ij}])||^2_2
	\end{aligned}
\end{equation}

The Multimodal Attentional Translation Embeddings (MATransE) model built upon VTransE\cite{gkanatsios2019deeply} learns a projection of $\left\langle \emph{S}, \emph{P}, \emph{O}\right\rangle$ into a score space where $S + P \approx O$, by guiding the features’ projection with attention to satisfy: 

\begin{equation}\label{eq:MATransE_Net}
\begin{aligned}
\bm{W}_p(s,o,m) x_p = \bm{W}_o(s,o,m) x_o - \bm{W}_s(s,o,m) x_s 
\end{aligned}
\end{equation}
where $x_+$ are the visual appearance features and $\bm{W}_+(s,o,m)$ are the projection matrices that are learned by employing a Spatio-Linguistic Attention module (SLA-M) that uses binary masks’ convolutional features $m$ and encodes subject and object classes with pre-trained word embeddings $s$, $o$. Compared with Eq. \ref{eq:VTransE_Net}, Eq. \ref{eq:MATransE_Net} can be interpreted as:

\begin{equation}\label{eq:MATransE_Net_interpreted}
\begin{aligned}
\bm{W}_o(s,o,m) x_o - \bm{W}_s(s,o,m) x_s = \bm{t}_p = \bm{W}_p(s,o,m) x_p
\end{aligned}
\end{equation}
Therefore, there are two branches, P-branch and OS-branch, to learn the relation translation vector $\bm{t}_p$ separately. To satisfy Eq. \ref{eq:MATransE_Net}, it enforces a score-level alignment by jointly minimizing the loss of each one of the P- and OS-branches with respect to ground-truth using deep supervision. 

Another TransE-inspired model is RLSV (Representation Learning via Jointly Structural and Visual Embedding)\cite{wan2018representation}. The architecture of RLSV is a three-layered hierarchical projection that projects a visual triple onto the attribute space, the relation space, and the visual space in order. This makes the \emph{subject} and \emph{object}, which are packed with attributes, projected onto the same space of the \emph{relation}, instantiated, and translated by the relation vector. This also makes the head entity and the tail entity packed with attributes, projected onto the same space of the relation, instantiated, and translated by the relation vector. It jointly combines the structural embeddings and the visual embeddings of a visual triple $t = (s, r, o)$ as new representations $(x_s, x_r, x_o)$ and scores it as follow:

\begin{equation}\label{eq:RLSV_score}
\begin{aligned}
E_I(t) = ||x_s + x_r - x_o||_{L1 / L2}
\end{aligned}
\end{equation}

~\\
\textcolor{black}{\textbf{In summary}, while many of the above 2D SGG models use more than one method, we selected the method we felt best reflected the idea of the paper for our primary classification of methods. The aforementioned 2D SGG challenges can also be addressed in other ways utilizing different concepts. As an example, Knyazev \emph{et al}.\cite{knyazev2020generative} used Generative Adversarial Networks (GANs) to synthesize rare yet plausible scene graphs to overcome the long-tailed distribution problem. Huang \emph{et al}.\cite{huang2020addressing} designed a Contrasting Cross-Entropy loss and a scoring module to address class imbalance. Fukuzawa and Toshiyuki\cite{fukuzawa2018problem} introduced a pioneering approach to visual relationship detection by reducing it to an object detection problem, and they won the Google AI Open Images V4 Visual Relationship Track Challenge. A neural tensor network was proposed by Qiang \emph{et al}.\cite{qiang2020tensor} for predicting visual relationships in an image. These methods contribute to the 2D SGG field in their own ways. }

\subsection{Spatio-Temporal Scene Graph Generation}

\begin{figure}[h] 
	\centering 
	\includegraphics[width=0.49\textwidth]{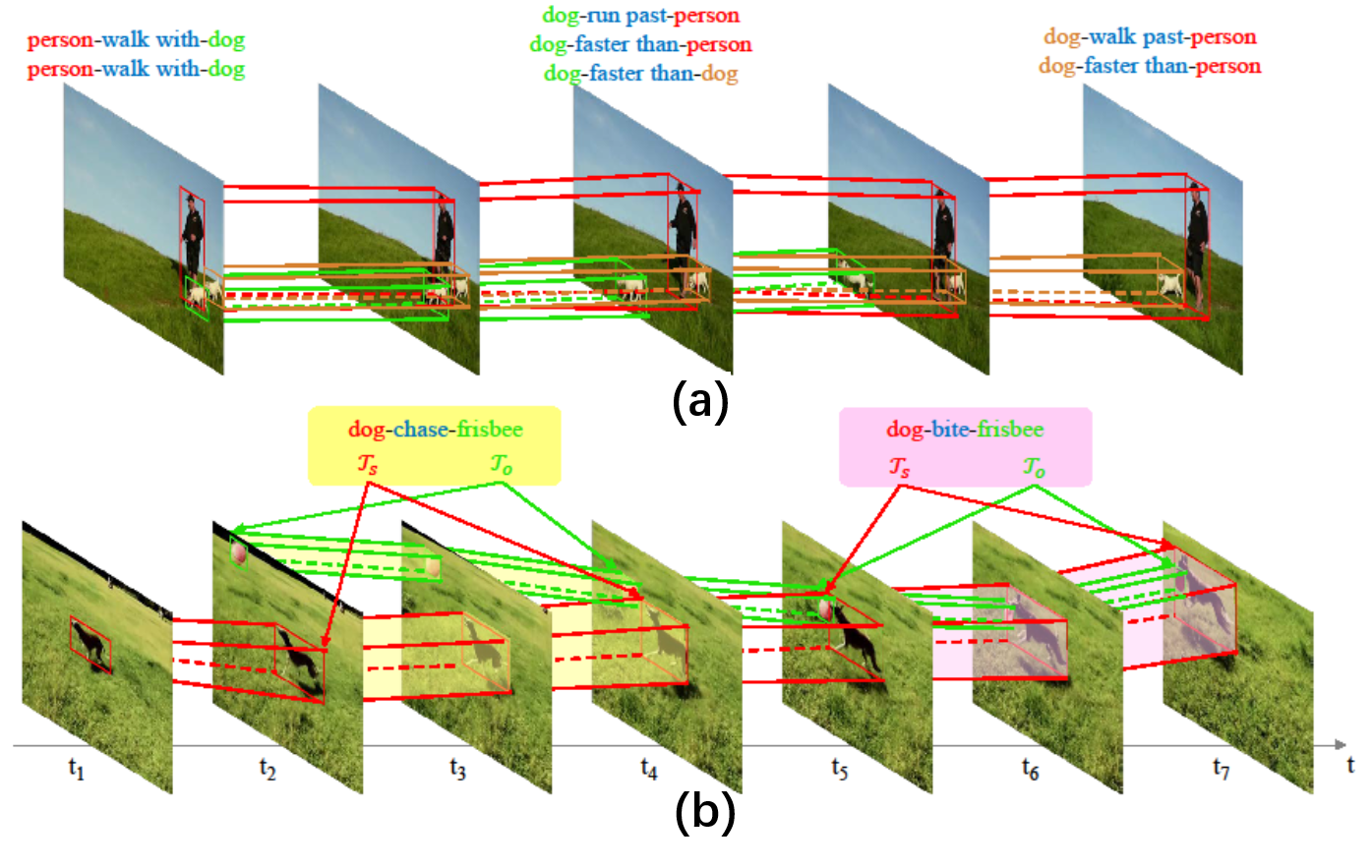} 
	\caption{Examples of video visual relations. (From \cite{shang2017video}).}
	\label{fig:VidVRD_Examples}  
\end{figure}

Recently, with the development of relationship detection models in the context of still images (ImgVRD), some researchers have started to pay attention to understand visual relationships in videos (VidVRD). Compared to images, videos provide a more natural set of features for detecting visual relations, such as the dynamic interactions between objects. Due to their temporal nature, videos enable us to model and reason about a more comprehensive set of visual relationships, such as those requiring temporal observations (e.g., {man, lift up, box} vs. {man, put down, box}), as well as relationships that are often correlated through time (e.g., {woman, pay, money} followed by {woman, buy, coffee}). Meanwhile, motion features extracted from spatial-temporal content in videos help to disambiguate similar predicates, such as ``walk" and ``run" ({Fig. \ref{fig:VidVRD_Examples}a}). Another significant difference between VidVRD and ImgVRD is that the visual relations in a video are usually changeable over time, while these of images are fixed. For instance, the objects may be occluded or out of one or more frame temporarily, which causes the occurrence and disappearance of visual relations. Even when two objects consistently appear in the same video frames, the interactions between them may be temporally changed\cite{shang2017video}. {Fig. \ref{fig:VidVRD_Examples}b} shows an example of temporally changing visual relation between two objects within a video. Two visual relation instances containing their relationship triplets and object trajectories of the subjects and objects.

Different from static images and because of the additional temporal channel, dynamic relationships in videos are often correlated in both the spatial and temporal dimensions. All the relationships in a video can collectively form a spatial-temporal graph structure, as mentioned in \cite{qian2019video, tsai2019video, ji2020action, liu2020beyond}. Therefore, we redefine the VidVRD as \bfseries Spatio-Temporal Scene Graph Generation (ST-SGG)\mdseries. To be consistent with the definition of 2D scene graph, we also define a spatio-temporal scene graph as a set of visual triplets $\mathcal{R}_S$. However, for each $r_{S, i\rightarrow j} = (s_{S, i}, p_{S, i\rightarrow j}, o_{S, j}) \in \mathcal{R}_S$, $s_{S, i} = (l_{S,k_1}, \mathcal{T}_s)$ and $o_{S, j} = (l_{S,k_2}, \mathcal{T}_o)$ both have a trajectory (resp. $\mathcal{T}_s$ and $\mathcal{T}_o$) rather than a fixed bbox. Specifically, $\mathcal{T}_s$ and $\mathcal{T}_o$ are two sequences of bounding boxes, which respectively enclose the subject and object, within the maximal duration of the visual relation. Therefore, VidVRD aims to detect each entire visual relation $\mathcal{R}_S$ instance with one bounding box trajectory.

ST-SGG relies on video object detection (VOD). The mainstream methods address VOD by integrating the latest techniques in both image-based object detection and multi-object tracking\cite{xiao2018video, shvets2019leveraging, wu2019sequence}. Although recent sophisticated deep neural networks have achieved superior performances in image object detection\cite{ren2015faster, redmon2018yolov3, cai2018cascade, zhou2019objects}, object detection in videos still suffers from a low accuracy, because of the presence of blur, camera motion and occlusion in videos, which hamper an accurate object localization with bounding box trajectories. Inevitably, these problems have gone down to downstream video relationship detection and even are amplified.

Shang \emph{et al}.\cite{shang2017video} first proposed VidVRD task and introduced a basic pipeline solution, which adopts a bottom-up strategy. The following models almost always use this pipeline, which decomposes the VidVRD task into three independent parts: multi-object tracking, relation prediction, and relation instances association. They firstly split videos into segments with a fixed duration and predict visual relations between co-occurrent short-term object tracklets for each video segment. Then they generate complete relation instances by a greedy associating procedure. Their object tracklet proposal is implemented based on a video object detection method similar to \cite{kang2016object} on each video segment. The relation prediction process consists of two steps: relationship feature extraction and relationship modeling. Given a pair of object tracklet proposals $(\mathcal{T}_s, \mathcal{T}_o)$ in a segment, \bfseries (1) \mdseries extract the improved dense trajectory (iDT) features\cite{wang2013action} with HoG, HoF and MBH in video segments, which capture both the motion and the low-level visual characteristics; \bfseries (2) \mdseries extract the relative characteristics between $\mathcal{T}_s$ and $\mathcal{T}_o$ which describes the relative position, size and motion between the two objects; \bfseries (3) \mdseries add the \bfseries classeme \mdseries feature\cite{zhang2017visual}. The concatenation of these three types of features as the overall relationship feature vector is fed into three predictors to classify the observed relation triplets. The dominating way to get the final video-level relationships is greedy local association, which greedily merges two adjacent segments if they contain the same relation.

Tsai \emph{et al}.\cite{tsai2019video} proposed a Gated Spatio-Temporal Energy Graph (GSTEG) that models the spatial and temporal structure of relationship entities in a video by a spatial-temporal fully-connected graph, where each node represents an entity and each edge denotes the statistical dependencies between the connected nodes. It also utilizes an energy function with adaptive parameterization to meet the diversity of relations, and achieves the state-of-the-art performance. The construction of the graph is realized by linking all segments as a Markov Random Fields (MRF) conditioned on a global observation. 

Shang \emph{et al}.\cite{shang2019annotating} has published another dataset, VidOR and launched the ACM MM 2019 Video Relation Understanding (VRU) Challenge$\footnote{https://videorelation.nextcenter.org/mm19-gdc/}$ to encourage researchers to explore visual relationships from a video\cite{shang2019relation}. In this challenge, Zheng \emph{et al}.\cite{zheng2019relation} use Deep Structural Ranking (DSR)\cite{liang2018visual} model to predict relations. Different from the pipeline in \cite{shang2017video}, they associate the short-term preliminary trajectories before relation prediction by using a sliding window method to locate the endpoint frames of a relationship triplet, rather than relational association at the end. Similarly, Sun \emph{et al}.\cite{sun2019video} also associate the preliminary trajectories on the front by applying a kernelized correlation filter (KCF) tracker to extend the preliminary trajectories generated by Seq-NMS in a concurrent way and generate complete object trajectories to further associate the short-term ones.

\subsection{3D Scene Graph Generation}

The classic computer vision methods aim to recognize objects and scenes in static images with the use of a mathematical model or statistical learning, and then progress to do motion recognition, target tracking, action recognition etc. in video. The ultimate goal is to be able to accurately obtain the shapes, positions and attributes of the objects in the three-dimensional space, so as to realize detection, recognition, tracking and interaction of objects in the real world. In the computer vision field, one of the most important branches of 3D research is the representation of 3D information. The common 3D representations are multiple views, point clouds, polygonal meshes, wireframe meshes and voxels of various resolutions. To extend the concept of scene graph to 3D space, researchers are trying to design a structured text representation to encode 3D information. Although existing scene graph research concentrates on 2D static scenes, based on these findings as well as on the development of 3D object detection\cite{Yang2018PIXORR3, Ali2018YOLO3DER, Shi2018PointRCNN3O, shi2020points} and 3D Semantic Scene Segmentation\cite{Qi2017PointNetDL,  Rethage2018FullyConvolutionalPN, Jaritz2019xMUDACU}, scene graphs in 3D have recently started to gain more popularity\cite{stuart2018visual, armeni20193d, kim20193-d, Wald2020Learning3S}. Compared with the 2D scene graph generation problem at the image level, to understand and represent the interaction of objects in the three-dimensional space is usually more complicated.

Stuart \emph{et al}.\cite{stuart2018visual} were the first to introduce the term of ``3D scene graph" and defined the problem and the model related to the prediction of 3D scene graph representations across multiple views. However, there is no essential difference in structure between their 3D scene graph and a 2D scene graph. Moreover, Zhang \emph{et al}.\cite{zhang2019support} started with the \emph{cardinal direction relations} and analyzed support relations between a group of connected objects grounded in a set of RGB-D images about the same static scene from different views. Kim \emph{et al}.\cite{kim20193-d} proposed a 3D scene graph model for robotics, which however takes the traditional scene graph only as a sparse and semantic representation of  three-dimensional physical environments for intelligent agents. To be precise, they just use the scene graph for the 3D scene understanding task. Similarly, Johanna \emph{et al}.\cite{Wald2020Learning3S} tried to understand indoor reconstructions by constructing 3D semantic scene graph. None of these works have proposed an ideal way to model the 3D space and multi-level semantics. 

Until now, there is no unified definition and representation of 3D scene graph. However, as an extension of the 2D scene graph in 3D spaces, 3D scene graph should be designed as a simple structure which encodes the relevant semantics within environments in an accurate, applicable, usable, and scalable way, such as object categories and relations between objects as well as physical attributes. It is noteworthy that, Armeni \emph{et al}.\cite{armeni20193d} creatively proposed a novel 3D scene graph model, which performs a hierarchical mapping of 3D models of large spaces in four stages: camera, object, room and building, and describe a semi-automatic algorithm to build the scene graph. Recently, Rosinol\emph{et al}.\cite{rosinol20203d} defined 3D Dynamic Scene Graphs as a unified representation for actionable spatial perception. More formally, this 3D scene graph is a layered directed graph where nodes represent spatial concepts (e.g., objects, rooms, agents) and edges represent pair-wise spatio-temporal relations (e.g., “agent $A$ is in room $B$ at time $t$”). They provide an example of a single-layer indoor environment which includes 5 layers (from low to high abstraction level): Metric-Semantic Mesh, Objects and Agents, Places and Structures, Rooms, and Building. Whether it is a four\cite{armeni20193d} - or five-story\cite{rosinol20203d} structure, we can get a hint that 3D scene contains rich semantic information that goes far beyond the 2D scene graph representation.

~\\
\textcolor{black}{\textbf{In summary}, this section includes a comprehensive overview of 2D SGG, followed by reviews of ST-SGG and 3D SGG. Researchers have contributed to the SGG field and will continue to do so, but the long-tail problem and the large intra-class diversity problem will remain hot issues, motivating researchers to explore more models to generate more useful scene graphs.}

\section{Datasets}
\label{sec:datasets}

In this section, we provide a summary of some of the most widely used datasets for visual relationship and scene graph generation. These datasets are grouped into three categories--2D images, videos and 3D representation.

\subsection{2D Datasets}

The majority of the research on visual relationship detection and scene graph generation has focused on 2D images; therefore, several 2D image datasets are available and their statistics are summarized in {Table \ref{table:2D-image-datasets}}. The following are some of the most popular ones:

\begin{table*}[ht]
	\renewcommand\arraystretch{1.4}
	\newcommand{\tabincell}[2]{\begin{tabular}{@{}#1@{}}#2\end{tabular}} 
	\centering
	\caption{The statistics of common 2D datasets.}
	\label{table:2D-image-datasets}
	\resizebox{\textwidth}{!}{
		\begin{tabular}{|l|r|r|r|r|r|c|} 
			\hline 
			Dataset & \makecell[c]{\emph{object}} & \makecell[c]{\emph{bbox}} & \makecell[c]{\emph{relationship}} & \makecell[c]{\emph{triplet}} & \makecell[c]{\emph{image}} & \emph{source link}\\
			\hline  
			Visual Phrase\cite{sadeghi2011recognition} & 8 & 3,271 & 9 & 1,796 & 2,769 & http://vision.cs.uiuc.edu/phrasal/\\
			Scene Graph\cite{Johnson2015Image} & 266 & 69,009 & 68 & 109,535 & 5,000 & \tabincell{c}{ http://imagenet.stanford.edu/internal/jcjohns/ \\ scene\_graphs/sg\_dataset.zip} \\
			VRD\cite{lu2016visual} & 100 & - & 70 & 37,993 & 5,000 & \tabincell{c}{https://cs.stanford.edu/people/ranjaykrishna/vrd/}\\
			Open Images v4\cite{kuznetsova2018open} & 57 & 3,290,070 & 329 & 374,768 & 9,178,275 & \tabincell{c}{https://storage.googleapis.com/openimages/ \\ web/index.html}\\
			Visual Genome\cite{krishna2017visual} & 33,877 & 3,843,636 & 40,480 & 2,347,187 & 108,077 & http://visualgenome.org/\\
			VrR-VG\cite{liang2019vrr-vg} & 1,600 & 282,460 & 117 & 203,375 & 58,983 & http://vrr-vg.com/\\
			UnRel\cite{zareian2020weakly} & - & - & 18 & 76 & 1,071 & https://www.di.ens.fr/willow/research/unrel/\\
			SpatialSense\cite{Yang2019SpatialSenseAA} & 3,679 & - & 9 & 13,229 & 11,569 & https://github.com/princeton-vl/SpatialSense\\
			SpatialVOC2K\cite{Belz2018SpatialVOC2KAM} & 20 & 5775 & 34 & 9804 & 2,026 & https://github.com/muskata/SpatialVOC2K\\
			\hline 
		\end{tabular}
	}
\end{table*}

\bfseries Visual Phrase \mdseries\cite{sadeghi2011recognition} is on visual phrase recognition and detection. The dataset contains 8 object categories from Pascal VOC2008\cite{everingham2010the} and 17 visual phrases that are formed by either an interaction between objects or activities of single objects. There are 2,769 images including 822 negative samples and on average 120 images per category. A total of 5,067 bounding boxes (1,796 for visual phrases + 3,271 for objects) were manually marked. 

\bfseries Scene Graph \mdseries\cite{Johnson2015Image} is the first dataset of real-world scene graphs. The full dataset consists of 5,000 images selected from the intersection of the YFCC100m\cite{Thomee2015TheND} and Microsoft COCO\cite{Lin2014MicrosoftCC} datasets and each of which has a human-generated scene graph.

\bfseries Visual Relationship Detection (VRD) \mdseries\cite{lu2016visual} dataset intends to benchmark the scene graph generation task. It highlights the long-tailed distribution of infrequent relationships. The public benchmark based on this dataset uses 4,000 images for training and test on the remaining 1,000 images. The relations broadly fit into categories, such as action, verbal, spatial, preposition and comparative.

\bfseries Visual Genome \mdseries\cite{krishna2017visual} has the maximum number of relation triplets with the most diverse object categories and relation labels up to now. Unlike VRD that is constructed by computer vision experts, VG is annotated by crowd workers and thus a substantial fraction of the object annotations has poor quality and overlapping bounding boxes and/or ambiguous object names. As an attempt to eliminate the noise, prior works have explored semi-automatic ways (e.g., class merging and filtering) to clean up object and relation annotations and constructed their own VG versions. Of these, VG200\cite{zhang2017visual}, VG150\cite{xu2017scene}, VG-MSDN\cite{li2017scene} and sVG\cite{dai2017detecting} have released their cleansed annotations and are the most frequently used. Other works\cite{cong2018scene, li2017vip, liang2017deep, plesse2018visual, yin2018zoom, yu2017visual, zhu2017visual, klawonn2018generating, zhou2019visual} use a paper-specific and nonpublicly available split, disabling direct future comparisons with their experiments. Moreover, \cite{zhang2019large} presents experiments on a large-scale version of VG, named VG80K, and \cite{liang2019vrr-vg} proposes a new split that has not been benchmarked yet. Sun\cite{sun2019hierarchical} constructed two datasets for hierarchical visual relationship detection (HVRD) based on VRD dataset and VG dataset, named H-VRD and H-VG, by expanding their flat relationship category spaces to hierarchical ones, respectively. The statistics of these datasets are summarized in {Table \ref{table:VG-versione-datasets}}.

\begin{table}[ht]
	\renewcommand\arraystretch{1.3}
	\centering
	\caption{The statistics of common VG versions.}
	\label{table:VG-versione-datasets}
	\resizebox{0.49\textwidth}{!}{
		\begin{tabular}{|l|r|r|r|r|r|} 
			\hline 
			Dataset & \makecell[c]{\emph{Pred. Classes}} & \makecell[c]{\emph{Obj. Classes}} & \makecell[c]{\emph{Total Images}} & \makecell[c]{\emph{Train Images}} & \makecell[c]{\emph{Test Images}}\\
			\hline  
			VG150\cite{xu2017scene} & 50 & 150 & 108,077 & 75.6k & 32.4k \\
			VG200\cite{zhang2017visual} & 100 & 200 & 99,658 & 73.8k & 25.8k \\
			sVG\cite{dai2017detecting} & 24 & 399 & 108,077 & 64.7k & 8.7k \\
			VG-MSDN\cite{li2017scene} & 50 & 150 & 95,998 & 71k & 25k \\
			VG80k\cite{zhang2019large} & 29,086 & 53,304 & 104,832 & 99.9k & 4.8k \\
			\hline 
		\end{tabular}
	}
\end{table}

\bfseries VG150 \mdseries\cite{xu2017scene} is constructed by pre-processing VG to improve the quality of object annotations. On average, this annotation refinement process has corrected 22 bounding boxes and/or names, deleted 7.4 boxes, and merged 5.4 duplicate bounding boxes per image. The benchmark uses the most frequent 150 object categories and 50 predicates for evaluation. As a result, each image has a scene graph of around 11.5 objects and 6.2 relationships. 

\bfseries VrR-VG \mdseries\cite{liang2019vrr-vg} is also based on Visual Genome. Its pre-processing aims at reducing the duplicate relationships by hierarchical clustering and filtering out the visually-irrelevant relationships. As a result, the dataset keeps the top 1,600 objects and 117 visually-relevant relationships of Visual Genome. Their hypothesis to identify visually-irrelevant relationships is that if a relationship label in different triplets is predictable according to any information, except visual information, the relationship is visually-irrelevant. This definition is a bit far-fetched but helps to eliminate redundant relationships.

\bfseries Open Images \mdseries\cite{kuznetsova2018open} is a dataset of ~9M images annotated with image-level labels, object bounding boxes, object segmentation masks, visual relationships, and localized narratives. The images are very diverse and often contain complex scenes with several objects (8.3 per image on average). It contains a total of 16M bounding boxes for 600 object classes on 1.9M images, making it the largest existing dataset with object location annotations. The boxes have largely been manually drawn by professional annotators to ensure accuracy and consistency. Open Images also offers visual relationship annotations, indicating pairs of objects in particular relations (e.g., ``woman playing guitar", ``beer on table"), object properties (e.g., ``table is wooden"), and human actions (e.g., ``woman is jumping"). In total it has 3.3M annotations from 1,466 distinct relationship triplets. So far, there are six released versions which are available on the official website and \cite{kuznetsova2018open} describes Open Images V4 in details, i.e., from the data collection and annotation to the detailed statistics about the data and the evaluation of the models trained on it.

\bfseries UnRel \mdseries\cite{zareian2020weakly} is a challenging dataset that contains 1,000 images collected from the web with 76 unusual language triplet queries such as “person ride giraffe”. All images are annotated at box-level for the given triplet queries. Since the triplet queries of UnRel are rare (and thus likely not seen at training), it is often used to evaluate the generalization performance of the algorithm.

\bfseries SpatialSense \mdseries\cite{Yang2019SpatialSenseAA} is a dataset specializing in spatial relation recognition. A key feature of the dataset is that it is constructed through \emph{adversarial crowdsourcing}: a human annotator is asked to come up with adversarial examples to confuse a recognition system. 

\bfseries SpatialVOC2K \mdseries\cite{Belz2018SpatialVOC2KAM} is the first multilingual image dataset with spatial relation annotations and object features for image-to-text generation. It consists of all 2,026 images with 9,804 unique object pairs from the PASCAL VOC2008 dataset. For each image, they provided additional annotations for each ordered object pair, i.e., (a) \emph{the single best}, and (b) \emph{all possible} prepositions that correctly describe the spatial relationship between objects. The preposition set contains 17 English prepositions and 17 French prepositions.

\subsection{Video Datasets}

The area of video relation understanding aims at promoting novel solutions and research on the topic of object detection, object tracking, action recognition, relation detection and spatio-temporal analysis, that are integral parts into a comprehensive visual system of the future. So far there are two public datasets for video relational understanding.

\bfseries ImageNet-VidVRD \mdseries\cite{shang2017video} is the first video visual relation detection dataset, which is constructed by selecting 1,000 videos from the training set and the validation set of ILSVRC2016-VID\cite{Russakovsky2015ImageNetLS}. Based on the 1,000 videos, the object categories increase to 35. It contains a total of 3,219 relationship triplets (\emph{i.e.,} the number of visual relation types) with 132 predicate categories. All videos were decomposed into segments of 30 frames with 15 overlapping frames in advance, and all the predicates appearing in each segment were labeled to obtain segment-level visual relation instances.

\bfseries VidOR \mdseries\cite{shang2019annotating} consists of 10,000 user-generated videos (98.6 hours) together with dense annotations on 80 categories of objects and 50 categories of predicates. The whole dataset is divided into 7,000 videos for training, 835 videos for validation, and 2,165 videos for testing. All the annotated categories of objects and predicates appear in each of the train/val/test sets. Specifically, objects are annotated with a bounding-box trajectory to indicate their spatio-temporal locations in the videos; and relationships are temporally annotated with start and end frames. The videos were selected from YFCC-100M multimedia collection and the average length of the videos is about 35 seconds. The relations are divided into two types, spatial relations (8 categories) and action relations (42 categories) and the annotation method is different for the two types of relations.

\subsection{3D Datasets}

Three dimensional data is usually provided via multi-view images such as point clouds, meshes, or voxels. Recently, several 3D datasets related to scene graphs have been released to satisfy the needs of SGG study. 

\bfseries 3D Scene Graph \mdseries is constructed by annotated the Gibson Environment Database\cite{Xia2018GibsonER} using the automated 3D Scene Graph generation pipeline proposed in \cite{armeni20193d}. Gibson’s underlying database of spaces includes 572 full buildings composed of 1,447 floors covering a total area of $211k m^2$. It is collected from real indoor spaces using 3D scanning and reconstruction and provides the corresponding 3D mesh model of each building. Meanwhile, for each space, the RGB images, depth and surface normals are provided. A fraction of the spaces is annotated with semantic objects.

\bfseries 3DSGG\mdseries, proposed in \cite{kim20193-d}, is a large scale 3D dataset that extends 3RScan with semantic scene graph annotations, containing relationships, attributes and class hierarchies. A scene graph here is a set of tuples $(N, R)$ between nodes $N$ and edges $R$. Each node is defined by a hierarchy of classes $c = (c_1, \cdots, c_d)$ and a set of attributes $A$ that describe the visual and physical appearance of the object instance. The edges define the semantic relations between the nodes. This representation shows that a 3D scene graph can easily be rendered to 2D.

\section{Performance Evaluation} 

In this section, we first introduce some commonly used evaluation modes and criteria for the scene graph generation task. Then, we provide the quantitative performance of the promising models on popular datasets. Since there is no uniform definition of a 3D scene graph, we will introduce these contents around 2D scene graph and spatio-temporal scene graph.

\subsection{Tasks} 
\label{sec:tasks_setup}

Given an image, the scene graph generation task consists of localizing a set of objects, classifying their category labels, and predicting relations between each pair of these objects. Most prior works often evaluated their SGG models on several of the following common sub-tasks. We preserve the names of tasks as defined in \cite{lu2016visual} and \cite{xu2017scene} here, despite the inconsistent terms used in other papers and the inconsistencies on whether they are in fact classification or detection tasks.

\begin{enumerate}

\item \bfseries Phrase Detection (PhrDet)\mdseries \cite{lu2016visual}: Outputs a label \emph{subject-predicate-object} and localizes the entire relationship in one bounding box with at least 0.5 overlap with the ground truth box. It is also called \bfseries Union boxes detection\mdseries in \cite{dai2017detecting}.

\item \bfseries Predicate Classification (PredCls) \mdseries \cite{xu2017scene}: Given a set of localized objects with category labels, decide which pairs interact and classify each pair’s predicate.

\item \bfseries Scene Graph Classification (SGCls) \mdseries \cite{xu2017scene}: Given a set of localized objects, predict the predicate as well as the object categories of the subject and the object in every pairwise relationship.

\item \bfseries Scene Graph Generation (SGGen) \mdseries \cite{xu2017scene}: Detect a set of objects and predict the predicate between each pair of the detected objects. This task is also called \bfseries Relationship Detection (RelDet) \mdseries in \cite{lu2016visual} or \bfseries Two boxes detection \mdseries in \cite{dai2017detecting}. It is similar to phrase detection, but with the difference that both the bounding box of the \emph{subject} and \emph{object} need at least 50 percent of overlap with their ground truth. Since SGGen only scores a single complete triplet, the result cannot reflect the detection effects of each component in the whole scene graph. So Yang \emph{et al}.\cite{yang2018graph} proposed the \bfseries Comprehensive Scene Graph Generation (SGGen+) \mdseries as an augmentation of SGGen. SGGen+ not only considers the triplets in the graph, but also the singletons (object and predicate). To be clear, SGGen+ is essentially a metric rather than a task.

\end{enumerate}

There are also some paper-specific task settings including \bfseries Triple Detection \mdseries \cite{baier2017improving}, \bfseries Relation Retrieval \mdseries \cite{zhang2017visual} and so on.

In the video based visual relationship detection task, there are two standard evaluation modes: \bfseries Relation Detection \mdseries and \bfseries Relation Tagging\mdseries. The detection task aims to generate a set of relationship triplets with tracklet proposals from a given video, while the tagging task only considers the accuracy of the predicted video relation triplets and ignores the object localization results. 

\subsection{Metrics} 

\bfseries Recall@K\mdseries. The conventional metric for the evaluation of SGG is the image-level $\bm{Recall@K (R@K)}$, which computes the fraction of times the correct relationship is predicted in the top $K$ confident relationship predictions. In addition to the most commonly used $R@50$ and $R@100$, some works also use the more challenging $R@20$ for a more comprehensive evaluation. Some methods compute R@K with the constraint that merely one relationship can be obtained for a given object pair. Some other works omit this constraint so that multiple relationships can be obtained, leading to higher values. There is a superparameter $k$, often not clearly stated in some works, which measures the maximum predictions allowed per object pair. Most works have seen PhrDet as a multiclass problem and they use $k = 1$ to reward the correct top-1 prediction for each pair. While other works\cite{liao2019natural, plesse2018learning, zheng2019visual} tackle this as a multilabel problem and they use a $k$ equal to the number of predicate classes to allow for predicate co-occurrences\cite{gkanatsios2019attention}. Some works\cite{gkanatsios2019attention, Lv2020AVRAB, zhang2019graphical, zellers2018neural, yu2017visual} have also identified this inconsistency and interpret it as whether there is graph constraint (\emph{i.e.,} the $k$ is the maximum number of edges allowed between a pair of object nodes). The unconstrained metric (i.e., no graph constraint) evaluates models more reliably, since it does not require a perfect triplet match to be the top-1 prediction, which is an unreasonable expectation in a dataset with plenty of synonyms and mislabeled annotations. For example, ‘man \emph{wearing} shirt’ and ‘man \emph{in} shirt’ are similar predictions, however, only the unconstrained metric allows for both to be included in ranking. Obviously, the SGGen+ metric above has a similar motivation as removing the graph constraint. Gkanatsios \emph{et al}.\cite{gkanatsios2019attention} re-formulated the metric as $\bm{Recall_k@K (R_k@K)}$. $k = 1$ is equivalent to ‘graph constraints” and a larger $k$ to “no graph constraints”, also expressed as $\bm{ngR_k@K}$. For $n$ examined subject-object pairs in an image, $\bm{Recall_k@K (R_k@K)}$ keeps the top-k predictions per pair and examines the $K$ most confident out of $nk$ total.

Given a set of ground truth triplets, $GT$, the image-level $\bm{R@K}$ is computed as:
\begin{equation}\label{eq:Recall_Compute}
\begin{aligned}
R@K = |Top_K \cap GT|\ /\ |GT|,
\end{aligned}
\end{equation}
where $Top_K$ is the top-K triplets extracted from the entire image based on ranked predictions of a model\cite{Knyazev2020GraphDL}. However, in the PredCLs setting, which is actually a simple classification task, the $\bm{R@K}$ degenerates into the triplet-level Recall@K ($\bm{R_{tr}@K}$). $\bm{R_{tr}@K}$ is similar to the top-K accuracy. Furthermore, Knyazev \emph{et al}.\cite{Knyazev2020GraphDL} proposed weighted triplet Recall($\bm{wR_{tr}@K}$), which computes a recall at each triplet and reweights the average result based on the frequency of the $GT$ in the training set:
\begin{equation}\label{eq:w-tr-Recall_Compute}
\begin{aligned}
wR_{tr}@K = \sum_t^T w_t[rank_t \leq K],
\end{aligned}
\end{equation}
where $T$ is the number of all test triplets, $[\cdot]$ is the Iverson bracket, $w_t = \frac {1}{(n_t+1) \sum_t 1 / (n_t+1) \in [0, 1]}$ and $n_t$ is the number of occurrences of the $t$-th triplet in the training set. It is friendly to those infrequent instances, since frequent triplets (with high $n_t$) are downweighted proportionally. To speak for all predicates rather than very few trivial ones, Tang \emph{et al}.\cite{tang2019learning} and Chen \emph{et al}.\cite{chen2019knowledge} proposed $\bm{mean Recall@K (mR@K)}$ which retrieves each predicate separately then averages $\bm{R@K}$ for all predicates.

Notably, there is an inconsistency in Recall’s definition on the entire test set: whether it is a \emph{micro-} or \emph{macro-Recall}\cite{gkanatsios2019attention}. Let $N$ be the number of testing images and $GT_i$ the ground-truth relationship annotations in image $i$. Then, having detected $TP_i = Top_{Ki} \cap GT_i$ true positives in the image $i$, micro-Recall micro-averages these positives as $\frac{\sum_i^N |TP_i|}{\sum_i^N |GT_i|}$ to reward correct predictions across dataset. Macro-Recall computed as $\frac{1}{N} \sum_i^N \frac{|TP_i|}{|GT_i|}$ macro-averages the detections in terms of images. Early works use micro-Recall on VRD and macro-Recall on VG150, but later works often use the two types interchangeably and without consistency. 

\noindent \bfseries Zero-Shot Recall@K\mdseries. Zero-shot relationship learning was proposed by Lu \emph{et al.}\cite{lu2016visual} to evaluate the performance of detecting zero-shot relationships. Due to the long-tailed relationship distribution in the real world, it is a practical setting to evaluate the extensibility of a model since it is difficult to build a dataset with every possible relationship. Besides, a single $\bm{wR_{tr}@K}$ value can show zero or few-shot performance linearly aggregated for all $n \geq 0$.

\noindent\bfseries Precision@K\mdseries. In the video relation detection task, $\bm{Precision@K (P@K)}$ is used to measure the accuracy of the tagging results for the relation tagging task.

\noindent\bfseries mAP\mdseries. In the OpenImages VRD Challenge, results are evaluated by calculating $Recall@50 (R@50)$, mean AP of relationships ($mAP_{rel}$), and mean AP of phrases ($mAP_{phr}$)\cite{zhang2019graphical}. The $mAP_{rel}$ evaluates AP of $\left\langle s, p, o\right\rangle$ triplets where both the subject and object boxes have an IOU of at least 0.5 with the ground truth. The $mAP_{phr}$ is similar, but applied to the enclosing relationship box. mAP would penalize the prediction if that particular ground truth annotation does not exist. Therefore, it is a strict metric because we can't exhaustively annotate all possible relationships in an image.

\subsection{Quantitative Performance}

\begin{table}[ht]
	\renewcommand\arraystretch{1.3}
	\newcommand{\tabincell}[2]{\begin{tabular}{@{}#1@{}}#2\end{tabular}}
	\centering
	\caption{Performance summary of some representative methods on VRD dataset.}
	\label{table:VRD-Performance}
	\resizebox{0.49\textwidth}{!}{
		\begin{tabular}{l|r|r|r|r|r|r|m{0.7cm}<{\centering}} 
			\hline 
			\multirow{2}*{\makecell[r]{Task \\ Models \qquad \qquad Metric}} &  \multicolumn{2}{c|}{PredCLs} & \multicolumn{2}{c|}{PhrDet} & \multicolumn{2}{c|}{RelDet} & \multirow{2}*{year}\\
			\cline{2-7}
			\multicolumn{1}{c|}{~} & R@100 & R@50 & R@100 & R@50 & R@100 & R@50\\
			\hline 
			LP\cite{lu2016visual} & 47.87 & 47.84 & 17.03 & 16.17 & 14.70 & 13.86 & 2016\\
			VRL\cite{liang2017deep} & - & - & 22.60 & 21.37 & 20.79 & 18.19 & 2017\\
			U+W+SF+L:S+T\cite{yu2017visual} & 55.16 & 55.16 & 24.03 & 23.14 & 21.34 & 19.17 & 2017\\
			DR-Net\cite{dai2017detecting} & 81.90 & 80.78 & 23.45 & 19.93 & 20.88 & 17.73 & 2017\\
			ViP-CNN\cite{li2017vip} & - & - & 27.91 & 22.78 & 20.01 & 17.32 & 2017\\
			AP+C+CAT\cite{zhuang2017towards} & 53.59 & 53.59 & 25.56 & 24.04 & 23.52 & 20.35 & 2017\\
			VTransE\cite{zhang2017visual} & - & - & 22.42 & 19.42 & 15.20 & 14.07 & 2017\\
			Cues\cite{Plummer2017PhraseLA} & - & - & 20.70 & 16.89 & 18.37 & 15.08 & 2017\\
			Weakly-supervised\cite{peyre2017weakly-supervised} & - & 46.80 & - & 16.00 & - & 14.10 & 2017\\
			PPR-FCN\cite{zhang2017ppr-fcn} & 47.43 & 47.43 & 23.15 & 19.62 & 15.72 & 14.41 & 2017\\
			Large VRU\cite{zhang2019large} & - & - & 39.66 & 32.90 & 32.63 & 26.98 & 2018\\
			Interpretable SGG\cite{zhang2018interpretable} & - & - & 41.25 & 33.29 & 32.55 & 26.67 & 2018\\
			CDDN-VRD\cite{Cui2018ContextDependentDN} & 93.76 & 87.57 & - & - & 26.14 & 21.46 & 2018\\
			DSR\cite{liang2018visual} & 93.18 & 86.01 & - & - & 23.29 & 19.03 & 2018\\
			Joint VSE\cite{li2018visual} & - & - & 24.12 & 20.53 & 16.26 & 14.23 & 2018\\
			F$_o$+L$^m$\cite{zhu2018deep} & - & - & 23.95 & 22.67 & 18.33 & 17.40 & 2018\\
			SG-CRF\cite{cong2018scene} & 50.47 & 49.16 & - & - & 25.48 & 24.98 & 2018\\
			OSL\cite{zhou2018object} & 56.56 & 56.56 & 24.50 & 20.82 & 16.01 & 13.81 & 2018\\
			F-Net\cite{li2018factorizable} & - & - & 30.77 & 26.03 & 21.20 & 18.32 & 2018\\
			Zoom-Net\cite{yin2018zoom} & 50.69 & 50.69 & 28.09 & 24.82 & 21.41 & 18.92 & 2018\\
			CAI+SCA-M\cite{yin2018zoom} & 55.98 & 55.98 & 28.89 & 25.21 & 22.39 & 19.54 & 2018\\
			VSA-Net\cite{han2018visual} & 49.22 & 49.22 & 21.65 & 19.07 & 17.74 & 16.03 & 2018\\
			MF-URLN\cite{zhan2019exploring} & 58.20 & 58.20 & 36.10 & 31.50 & 26.80 & 23.90 & 2019\\
			LRNNTD\cite{dupty2020visual} & - & - & 30.92 & 28.53 & 25.87 & 24.20 & 2019\\
			KB-GAN\cite{gu2019scene} & - & - & 34.38 & 27.39 & 25.01 & 20.31 & 2019\\
			RLM\cite{zhou2019visual} & 57.19 & 57.19 & 39.74 & 33.20 & 31.15 & 26.55 & 2019\\
			NMP\cite{Hu2019NeuralMP} & 57.69 & 57.69 & - & - & 23.98 & 20.19 & 2019\\
			MLA-VRD\cite{zheng2019visual} & 95.05 & 90.18 & 28.12 & 23.36 & 24.91 & 20.54 & 2019\\
			ATR-Net\cite{gkanatsios2019attention} & 58.40 & 58.40 & 34.63 & 29.74 & 24.87 & 22.83 & 2019\\
			BLOCK\cite{benyounes2019block} & 92.58 & 86.58 & 28.96 & 26.32 & 20.96 & 19.06 & 2019\\
			MR-Net\cite{bin2019mr-net} & 61.19 & 61.19 & - & - & 17.58 & 16.71 & 2019\\
			RelDN\cite{zhang2019graphical} & - & - & 36.42 & 31.34 & 28.62 & 25.29 & 2019\\
			
			UVTransE\cite{hung2020contextual} & - & 26.49 & 18.44 & 13.07 & 16.78 & 11.00 & 2020\\
			AVR\cite{Lv2020AVRAB} & 55.61 & 55.61 & 33.27 & 29.33 & 25.41 & 22.83 & 2020\\
			GPS-Net\cite{Lin2020GPSNetGP} & - & 63.40 & 39.20 & 33.80 & 31.70 & 27.80 & 2020\\
			MemoryNet\cite{wang2020memory} & - & - & 34.90 & 29.80 & 27.90 & 24.30 & 2020\\
			HET\cite{wang2020sketching} & - & - & 42.94 & 35.47 & 24.88 & 22.42 & 2020\\
			HOSE-Net\cite{wei2020hose} & - & - & 31.71 & 37.04 & 23.57 & 20.46 & 2020\\
			SABRA\cite{jin2020towards} & - & - & 39.62 & 33.56 & 32.48 & 27.87 & 2020\\
			
			\hline
			NLGVRD$^\ddagger$\cite{liao2019natural} & 92.65 & 84.92 & 47.92 & 42.29 & 22.22 & 20.81 & 2017\\
			U+W+SF+L:S+T$^\ddagger$\cite{yu2017visual} & 94.65 & 85.64 & 29.43 & 26.32 & 31.89 & 22.68 & 2017\\
			Zoom-Net$^\ddagger$\cite{yin2018zoom} & 90.59 & 84.05 & 37.34 & 29.05 & 27.30 & 21.37 & 2018\\
			CAI+SCA-M$^\ddagger$\cite{yin2018zoom} & 94.56 & 89.03 & 38.39 & 29.64 & 28.52 & 22.34 & 2018\\
			LRNNTD$^\ddagger$\cite{dupty2020visual} & - & - & 41.28 & 32.29 & 34.93 & 27.09 & 2019\\
			RLM$^\ddagger$\cite{zhou2019visual} & 96.48 & 90.00 & 46.03 & 36.79 & 37.35 & 30.22 & 2019\\
			NMP$^\ddagger$\cite{Hu2019NeuralMP} & 96.61 & 90.61 & - & - & 27.50 & 21.50  & 2019\\
			ATR-Net$^\ddagger$\cite{gkanatsios2019attention} & 96.97 & 91.00 & 41.01 & 33.20 & 31.94 & 26.04 & 2019\\
			RelDN$^\ddagger$\cite{zhang2019graphical} & - & - & 42.12 & 34.45 & 33.91 & 28.15 & 2019\\
			AVR$^\ddagger$\cite{Lv2020AVRAB} & 95.72 & 90.73 & 41.36 & 34.51 & 32.96 & 27.35 & 2020\\
			MemoryNet$^\ddagger$\cite{wang2020memory} & - & - & 39.80 & 32.10 & 32.40 & 26.50 & 2020\\
			HET$^\ddagger$\cite{wang2020sketching} & - & - & 43.05 & 35.47 & 31.81 & 26.88 & 2020\\
			HOSE-Net$^\ddagger$\cite{wei2020hose} & - & - & 36.16 & 28.89 & 27.36 & 22.13 & 2020\\
			SABRA$^\ddagger$\cite{jin2020towards} & - & - & 45.29 & 36.62 & 37.71 & 30.71 & 2020\\
			\hline
		\end{tabular}
	}
\end{table}

\begin{table}[ht]
	\renewcommand\arraystretch{1.3}
	\newcommand{\tabincell}[2]{\begin{tabular}{@{}#1@{}}#2\end{tabular}}
	\centering
	\caption{Performation summary of some representative methods on VG150 dataset.}
	\label{table:VG150-Performance}
	\resizebox{0.49\textwidth}{!}{
		\begin{tabular}{l|r|r|r|r|r|r|m{0.7cm}<{\centering}} 
			\hline 
			\multirow{2}*{\makecell[r]{Task \\ Models \qquad \qquad Metric}} &  \multicolumn{2}{c|}{PredCLs} & \multicolumn{2}{c|}{PhrDet} & \multicolumn{2}{c|}{RelDet} & \multirow{2}*{year}\\
			\cline{2-7}
			\multicolumn{1}{c|}{~} & R@100 & R@50 & R@100 & R@50 & R@100 & R@50\\
			\hline 
			IMP\cite{xu2017scene} & 53.08 & 44.75 & 24.38 & 21.72 & 4.24 & 3.44 & 2017\\
			Px2graph\cite{newell2017pixels} & 86.40 & 82.00 & 38.40 & 35.70 & 18.80 & 15.50 & 2017\\
			Interpretable SGG\cite{zhang2018interpretable} & 68.30 & 68.30 & 36.70 & 36.70 & 32.50 & 28.10 & 2018\\
			IK-$R_e$\cite{plesse2018visual} & 77.60 & 67.71 & 42.74 & 35.55 & - & - & 2018\\
			SK-$R_e$\cite{plesse2018visual} & 77.43 & 67.42 & 42.25 & 35.07 & - & - & 2018\\
			TFR\cite{Hwang2018TensorizeFA} & 58.30 & 51.90 & 26.60 & 24.30 & 6.00 & 4.80 & 2018\\
			MotifNet\cite{zellers2018neural} & 67.10 & 65.20 & 36.50 & 35.80 & 30.30 & 27.20 & 2018\\
			Graph R-CNN\cite{yang2018graph} & 59.10 & 54.20 & 31.60 & 29.60 & 13.70 & 11.40 & 2018\\
			LinkNet\cite{woo2018linknet} & 68.50 & 67.00 & 41.70 & 41.00 & 30.10 & 27.40 & 2018\\
			GPI\cite{herzig2018mapping} & 66.90 & 65.10 & 38.80 & 36.50 & - & - & 2018\\
			KERN\cite{chen2019knowledge} & 67.60 & 65.80 & 37.40 & 36.70 & 29.80 & 27.1 & 2019\\
			SGRN\cite{liao2019exploring} & 66.40 & 64.20 & 39.70 & 38.60 & 35.40 & 32.30 & 2019\\
			Mem+Mix+Att\cite{wang2019exploring} & 57.90 & 53.20 & 29.50 & 27.80 & 13.90 & 11.40 & 2019\\
			VCTREE\cite{tang2019learning} & 68.10 & 66.40 & 38.80 & 38.10 & 31.30 & 27.90 & 2019\\
			CMAT\cite{chen2019counterfactual} & 68.10 & 66.40 & 39.80 & 39.00 & 31.20 & 27.90 & 2019\\
			VRasFunctions\cite{dornadula2019visual} & 57.21 & 56.65 & 24.66 & 23.71 & 13.45 & 13.18 & 2019\\
			PANet\cite{chen2019panet} & 67.90 & 66.00 & 41.80 & 40.90 & 29.90 & 26.90 & 2019\\
			ST+GSA+RI\cite{qi2019attentive} & 61.30 & 56.60 & 40.40 & 38.20 & - & - & 2019\\
			Attention\cite{zhang2019relationship} & 67.10 & 65.00 & 37.10 & 36.30 & 29.50 & 26.60 & 2019\\
			RelDN\cite{zhang2019graphical} & 68.40 & 68.40 & 36.80 & 36.80 & 32.70 & 28.30 & 2019\\
			Large VRU\cite{zhang2019large} & 68.40 & 68.40 & 36.70 & 36.70 & 32.50 & 27.9 & 2019\\
			GB-NET\cite{zareian2020bridging} & 68.20 & 66.60 & 38.80 & 38.00 & 30.00 & 26.40 & 2020\\
			UVTransE\cite{hung2020contextual} & 67.30 & 65.30 & 36.60 & 35.90 & 33.60 & 30.10 & 2020\\
			GPS-Net\cite{Lin2020GPSNetGP} & 69.70 & 69.70 & 42.30 & 42.30 & 33.20 & 28.90 & 2020\\
			RiFa\cite{wen2020unbiased} & 88.35 & 80.64 & 44.38 & 37.62 & 26.68 & 20.86 & 2020\\
			RONNIE\cite{kenigsfield2019leveraging} & 69.00 & 65.00 & 37.00 & 36.20 & - & - & 2020\\
			DG-PGNN\cite{Khademi2020DeepGP} & 73.00 & 70.10 & 40.80 & 39.50 & 33.10 & 32.10 & 2020\\
			NODIS\cite{yuren2020nodis} & 69.10 & 67.20 & 41.50 & 40.60 & 31.50 & 28.10 & 2020\\
			HCNet\cite{Ren2020SceneGG} & 68.80 & 66.40 & 37.30 & 36.60 & 31.20 & 28.00 & 2020\\
			Self-supervision\cite{inuganti2020assisting} & 68.87 & 68.85 & 37.03 & 37.01 & 32.56 & 28.28 & 2020\\
			MemoryNet\cite{wang2020memory} & 69.30 & 69.20 & 37.10 & 37.10 & 32.40 & 27.60 & 2020\\
			HET\cite{wang2020sketching} & 68.10 & 66.30 & 37.30 & 36.60 & 30.90 & 27.50 & 2020\\
			HOSE-Net\cite{wei2020hose} & 69.20 & 66.70 & 37.40 & 36.30 & 33.30 & 28.90 & 2020\\
			PAIL\cite{tian2020part} & 69.40 & 67.70 & 40.20 & 39.40 & 32.70 & 29.40 & 2020\\
			BL+SO+KT+FC\cite{he2020learning} & 68.80 & 66.20 & 38.30 & 37.50 & 31.40 & 28.20 & 2020\\
			
			\hline
			Interpretable SGG$^\ddagger$\cite{zhang2018interpretable} & 97.70 & 93.70 & 50.80 & 48.90 & 36.40 & 30.10 & 2018\\
			MotifNet$^\ddagger$\cite{zellers2018neural} & 88.30 & 81.10 & 47.70 & 44.50 & 35.80 & 30.50 & 2018\\
			GPI$^\ddagger$\cite{herzig2018mapping} & 88.20 & 80.80 & 50.80 & 45.50 & - & - & 2018\\
			KERN$^\ddagger$\cite{chen2019knowledge} & 88.90 & 81.90 & 49.00 & 45.90 & 35.80 & 30.90 & 2019\\
			CMAT$^\ddagger$\cite{chen2019counterfactual} & 90.10 & 83.20 & 52.00 & 48.60 & 36.80 & 31.60 & 2019\\
			PANet$^\ddagger$\cite{chen2019panet} & 89.70 & 82.60 & 55.20 & 51.30 & 36.30 & 31.10 & 2019\\
			RelDN$^\ddagger$\cite{zhang2019graphical} & 97.80 & 93.80 & 50.80 & 48.90 & 36.70 & 30.40 & 2019\\
			GB-NET$^\ddagger$\cite{zareian2020bridging} & 90.50 & 83.60 & 51.10 & 47.70 & 35.10 & 29.40 & 2020\\
			HOSE-Net$^\ddagger$\cite{wei2020hose} & 89.20 & 81.10 & 48.10 & 44.20 & 36.30 & 30.50 & 2020\\
			BL+SO+KT+FC$^\ddagger$\cite{he2020learning} & 90.20 & 82.50 & 50.20 & 46.20 & 36.50 & 31.40 & 2020\\
			\hline
			
		\end{tabular}
	}
\end{table}

We present the quantitative performance on Recall@K metric of some representative methods on several commonly used datasets in {Table \ref{table:VRD-Performance}-\ref{table:VG150-Performance}}. We preserve the respective task settings and tasks’ names for each dataset, though SGGen on VG150 are the same to the RelDet on others. $\ddagger$ denotes the experimental results are under \bfseries ``no graph constraints"\mdseries.

By comparing Table 3 and Table 4, we notice that only a few of the proposed methods have been simultaneously verified on both VRD and VG150 datasets. The performance of most methods on VG150 is better than that on VRD dataset, because VG150 has been cleaned and enhanced. Experimental results on VG150 can better reflect the performance of different methods, therefore, several recently proposed methods have adopted VG150 to compare their performance metrics with other techniques.

Recently, two novel techniques i.e., SABRA\cite{jin2020towards} and HET\cite{wang2020sketching} have achieved SOTA performance for PhrDet and RelDet on VRD, respectively. SABRA enhanced the robustness of the training process of the proposed model by subdividing negative samples, while HET followed the intuitive perspective i.e., the more salient the object, the more important it would be for the scene graph.

On VG150, excellent performances have been achieved by using the Language Prior's model, especially RiFa\cite{wen2020unbiased}. In particular, RiFa has achieved good results on the unbalanced data distribution by mining the deep semantic information of the objects and relations in triplets. SGRN\cite{liao2019exploring} generates the initial scene graph structure using the semantic information, to ensure that its information transmission process accepts the positive influence from the semantic information. Theoretically, Commonsense Knowledge can greatly improve the performance, but in practice, several models that use Prior Knowledge have unsatisfactory performance. We believe the main reason is the difficultly to extract and use the effective knowledge information in the scene graph generation model. Gb-net\cite{zareian2020bridging} has paid attention to this problem, and achieved good results in PredDet and PhrDet by establishing connection between scene graph and Knowledge Graph, which can effectively use the commonsense knowledge.

Due to the long tail effect of visual relationships, it is hard to collect images for all the possible relationships. It is therefore crucial for a model to have the generalizability to detect zero-shot relationships. VRD dataset contains 1,877 relationships that only exist in the test set. Some researchers have evaluated the performance of their models on zero-shot learning. The performance summary of zero-shot predicate and relationship detection on VRD dataset are shown in {Table \ref{table:VRD-Zero_Shot-Performance}}.

\begin{table}[ht]
	\renewcommand\arraystretch{1.3}
	\newcommand{\tabincell}[2]{\begin{tabular}{@{}#1@{}}#2\end{tabular}}
	\centering
	\caption{Performance summary of some representative methods for \bfseries zero-shot \mdseries visual relationship detection on the VRD dataset.}
	\label{table:VRD-Zero_Shot-Performance}
	\resizebox{0.49\textwidth}{!}{
		\begin{tabular}{l|r|r|r|r|r|r|m{0.7cm}<{\centering}} 
			\hline 
			\multirow{2}*{\makecell[r]{Task \\ Models \qquad \qquad Metric}} &  \multicolumn{2}{c|}{PredCLs} & \multicolumn{2}{c|}{PhrDet} & \multicolumn{2}{c|}{RelDet} & \multirow{2}*{year}\\
			\cline{2-7}
			\multicolumn{1}{c|}{~} & R@100 & R@50 & R@100 & R@50 & R@100 & R@50\\
			\hline 
			LP\cite{lu2016visual} & 8.45 & 8.45 & 3.75 & 3.36 & 3.52 & 3.13 & 2016\\
			VRL\cite{liang2017deep} & - & - & 10.31 & 9.17 & 8.52 & 7.94 & 2017\\
			Cues\cite{Plummer2017PhraseLA} & - & - & 15.23 & 10.86 & 13.43 & 9.67 & 2017\\
			VTransE\cite{zhang2017visual} & - & - & 3.51 & 2.65 & 2.14 & 1.71 & 2017\\
			Weakly-supervised\cite{peyre2017weakly-supervised} & - & 19.00 & - & 6.90 & - & 6.70 & 2017\\
			U+W+SF+L:S\cite{yu2017visual} & 16.98 & 16.98 & 10.89 & 10.44 & 9.14 & 8.89 & 2017\\
			AP+C+CAT\cite{zhuang2017towards} & 16.37 & 16.37 & 11.30 & 10.78 & 10.26 & 9.54 & 2017\\
			PPR-FCN\cite{zhang2017ppr-fcn} & - & - & 8.22 & 6.93 & 6.29 & 5.68 & 2017\\
			DSR\cite{liang2018visual} & 79.81 & 60.90 & - & - & 9.20 & 5.25 & 2018\\
			CDDN-VRD\cite{Cui2018ContextDependentDN} & 84.00 & 67.66 & - & - & 10.29 & 6.40 & 2018\\ 
			Joint VSE\cite{li2018visual} & - & - & 6.16 & 5.05 & 5.73 & 4.79 & 2018\\
			SG-CRF\cite{cong2018scene} & 21.22 & - & 6.70 & - & 5.22 & - & 2018\\
			MF-URLN\cite{zhan2019exploring} & 26.90 & 26.90 & 7.90 & 5.90 & 5.50 & 4.30 & 2019\\
			MLA-VRD\cite{zheng2019visual} & 88.96 & 73.65 & 13.84 & 8.43 & 12.81 & 8.08 & 2019\\
			\hline
			U+W+SF+L:S$^\ddagger$\cite{yu2017visual} & 74.65 & 54.20 & 17.24 & 13.01 & 16.15 & 12.31 & 2017\\
			\hline
			
		\end{tabular}
	}
\end{table}

Compared with the traditional Recall, meanRecall calculates a Recall rate for each relation. Therefore, meanRecall can better describe the performance of the model on each relation, which is obtained by averaging the Recall of each relation. {Table \ref{table:VG150-MeanPerformance}} shows the meanRecall metric performance of several typical models.  In {Table \ref{table:VG150-MeanPerformance}}, IMP's meanRecall performance in detecting tail relationships is not ideal. In IMP+, due to the introduction of bidirectional LSTM to extract the characteristics of each object, more attention has been paid to the object itself, so there is an improvement. The core idea of VCTREE comes from MotifNet, but it improves the strategy of information transmission by changing the chain structure to the tree structure, making the information transmission between objects more directional.  MemoryNet
\cite{wang2020memory} has achieved SOTA results on both PredCLs and SGGen, which focuses on semantic overlap between low and high frequency relationships. 

\begin{table}[ht]
	\renewcommand\arraystretch{1.3}
	\newcommand{\tabincell}[2]{\begin{tabular}{@{}#1@{}}#2\end{tabular}}
	\centering
	\caption{Mean Recall performance summary of some typical methods on  VG150 dataset.}
	\label{table:VG150-MeanPerformance}
	\resizebox{0.49\textwidth}{!}{
		\begin{tabular}{l|r|r|r|r|r|r|m{0.7cm}<{\centering}} 
			\hline 
			\multirow{2}*{\makecell[r]{Task \\ Models \qquad \qquad Metric}} &  \multicolumn{2}{c|}{PredCLs} & \multicolumn{2}{c|}{SGCLs} & \multicolumn{2}{c|}{SGGen} & \multirow{2}*{year}\\
			\cline{2-7}
			\multicolumn{1}{c|}{~} & mR@50 & mR@100 & mR@50 & mR@100 & mR@50 & mR@100\\
			\hline 
			IMP\cite{xu2017scene} & 6.1 & 8.0 & 3.1 & 3.8 & 0.6 & 0.9 & 2017\\
			IMP+\cite{zellers2018neural} & 9.8 & 10.5 & 5.8 & 6.0 & 3.8 & 4.8 & 2018\\
			FREQ\cite{zellers2018neural} & 13.0 & 16.0 & 7.2 & 8.5 & 6.1 & 7.1 & 2018\\
			MotifNet\cite{zellers2018neural} & 14.0 & 15.3 & 7.7 & 8.2 & 5.7 & 6.6 & 2018\\
			KERN\cite{chen2019knowledge} & 17.7 & 19.2 & 9.4 & 10 & 6.4 & 7.3 & 2019\\
			VCTREE-SL\cite{tang2019learning} & 17.0 & 18.5 & 9.8 & 10.5 & 6.7 & 7.7 & 2019\\
			VCTREE-HL\cite{tang2019learning} & 17.9 & 19.4 & 10.1 & 10.8 & 6.9 & 8 & 2019\\
			GPS-Net\cite{Lin2020GPSNetGP} & 21.3 & 22.8 & 11.8 & 12.6 & 8.7 & 9.8 & 2020\\
			MemoryNet\cite{wang2020memory} & 22.6 & 22.7 & 10.9 & 11 & 7.4 & 9 & 2020\\
			PAIL\cite{tian2020part} & 19.2 & 20.9 & 10.9 & 11.6 & 7.7 & 8.8 & 2020\\
			GB-NET\cite{zareian2020bridging} & 19.3 & 20.9 & 9.6 & 10.2 & 6.1 & 7.3 & 2020\\
			GB-NET-$\beta$\cite{zareian2020bridging} & 22.1 & 24.0 & 12.7 & 13.4 & 7.1 & 8.5 & 2020\\
			\hline
			
		\end{tabular}
	}
\end{table}

\begin{table}[ht]
	\renewcommand\arraystretch{1.3}
	\newcommand{\tabincell}[2]{\begin{tabular}{@{}#1@{}}#2\end{tabular}}
	\centering
	\caption{\textcolor{black}{Performance for standard video relation detection and video relation tagging on ImageNet-VidVRD dataset\cite{liu2020beyond}.}}
	\label{table:ImageNet-VidVRD}
	\resizebox{0.49\textwidth}{!}{
		\begin{tabular}{l|r|r|r|r|r|r|m{0.7cm}<{\centering}} 
			\hline 
			\multirow{2}*{\makecell[r]{Task \\ Models \qquad \qquad Metric}} &  \multicolumn{3}{c|}{Relation Detection} & \multicolumn{3}{c|}{Relation Tagging} &  \multirow{2}*{year}\\
			\cline{2-7}
			\multicolumn{1}{c|}{~} & R@50 & R@100 & mAP & P@1 & P@5 & P@10\\
			\hline 
			VP\cite{sadeghi2011recognition} & 0.89 & 1.41 & 1.01 & 36.50 & 25.55 & 19.20 & 2011\\
			Lu's-V\cite{lu2016visual} & 0.99 & 1.80 & 2.37 & 20.00 & 12.60 & 9.55 & 2016\\
			Lu's\cite{lu2016visual} & 1.10 & 2.23 & 2.40 & 20.50 & 16.30 & 14.05 & 2016\\
			VTransE\cite{zhang2017visual} & 0.72 & 1.45 & 1.23 & 15.00 & 10.00 & 7.65 & 2017\\
			VidVRD\cite{shang2017video} & 5.54 & 6.37 & 8.58 & 43.00 & 28.90 & 20.80 & 2017\\
			GSTEG\cite{tsai2019video} & 7.05 & 8.67 & 9.52 & 51.50 & 39.50 & 28.23 & 2019\\
			VRD-GCN\cite{qian2019video} & 8.07 & 9.33 & 16.26 & 57.50 & 41.00 & 28.50 & 2019\\
			VRD-STGC\cite{liu2020beyond} & 11.21 & 13.69 & 18.38 & 60.00 & 43.10 & 32.24 & 2020\\
			\hline
			
		\end{tabular}
	}
\end{table}

\begin{table}[ht]
	\renewcommand\arraystretch{1.3}
	\newcommand{\tabincell}[2]{\begin{tabular}{@{}#1@{}}#2\end{tabular}}
	\centering
	\caption{\textcolor{black}{Performance for standard video relation detection and video relation tagging on VidOR dataset\cite{liu2020beyond}.}}
	\label{table:VidOR}
	\resizebox{0.49\textwidth}{!}{
		\begin{tabular}{l|r|r|r|r|r|m{0.7cm}<{\centering}} 
			\hline 
			\multirow{2}*{\makecell[r]{Task \\ Models \qquad \qquad Metric}} &  \multicolumn{3}{c|}{Relation Detection} & \multicolumn{2}{c|}{Relation Tagging} &  \multirow{2}*{year}\\
			\cline{2-6}
			\multicolumn{1}{c|}{~} & R@50 & R@100 & mAP & P@1 & P@5\\
			\hline 
			RELAbuilder\cite{zheng2019relation} & 1.58 & 1.85 & 1.47 & 33.05 & 35.27 & 2019\\
			OTD+CAI\cite{sun2019video} & 6.19 & 8.16 & 5.65 & 48.31 & 38.49 & 2019\\
			OTD+GSTEG\cite{sun2019video} & 6.40 & 8.43 & 5.58 & 51.20 & 37.26 & 2019\\
			MAGUS.Gamma\cite{sun2019video} & 6.89 & 8.83 & 6.56 & 51.20 & 40.73 & 2019\\
			VRD-STGC\cite{liu2020beyond} & 8.21 & 9.90 & 6.85 & 48.92 & 36.78 & 2020\\
			\hline
			
		\end{tabular}
	}
\end{table}

\textcolor{black}{Tables \ref{table:ImageNet-VidVRD} and \ref{table:VidOR} show the performances of several ST-SGG methods on the ImageNet-VidVRD and VidOR datasets. Evaluation of performance is based on two tasks, namely Relation Detection and Relation Tagging. Because of its large size, VidOR presents many challenges to relation detection and tagging. ST-SGG is much more complex than 2D SGG because additional steps such as object tracking, temporal segmentation, and merging the detected relationships in different segments are involved. It is expected that ST-SGG's performance will improve as more researchers contribute.}

\textcolor{black}{A fair comparison between 3D SGG methods cannot currently be undertaken due to a lack of a unified definition of 3D scene graphs. With rapid advances in 3D object detection, segmentation and description, 3D SGG should be able to provide unified tasks, evaluation metrics, as well as quantitative performances in the near future.}

\section{Challenges \& Future Research Directions}

\subsection{Challenges} 

There is not doubt that there are many excellent SGG models which have achieved good performance on the standard image datasets, such as VRD and VG150. However, there are still several challenges that have not been well resolved. 

\textcolor{black}{\bfseries First\mdseries, both the number of objects in the real world and the number of categories of relations are very large, but reasonable and meaningful relationships are scarce. Therefore, detecting all individual objects first and then classifying all pairs would be inefficient. Moreover, classification requires a limited number of object categories, which does not scale with real-world images.} Several works\cite{zhang2017relationship, dai2017detecting, liao2019exploring, zhang2019large, yang2018graph, li2017vip, guo2020one, zhang2017ppr-fcn} have helped to filter out a set of object pairs with a low probability of interaction from the set of detected objects. An effective proposal network will definitely reduce the learning complexity and computational cost for the subsequent predicate classification, thus improving the accuracy of relationship detection. 

The \textbf{\emph{second}} main challenge comes from the long-tailed distribution of the visual relationships. Since interaction occurs between two objects, there is a greater skew of rare relationships, as object co-occurrence is infrequent in a real-world scenario. An uneven distribution makes it difficult for the model to fully understand the properties of some rare relationships and triplets. For example, if a model is trained to predict ‘on’ 1,000 times more than ‘standing on’, then, during the test phase, “on” is more likely to prevail over “standing on”. 
\textbf{\emph{This phenomenon where the model is more likely to predict a simple and coarse relation than the accurate relation is called Biased Scene Graph Generation. }}
Under this condition, even though the model can output a reasonable predicate, it is too coarse and obscure to describe the scene. However, for several downstream tasks, an accurate and informative pair-wise relation is undoubtedly the most fundamental requirement. Therefore, to perform a sensible graph reasoning, we need to distinguish between the more fine-grained relationships from the ostensibly probable but trivial ones, which is generally regarded as unbiased scene graph generation. 
A lot of works\cite{zhang2017visual, yu2017visual, Plummer2017PhraseLA, liang2018visual, cong2018scene, zheng2019visual, Cui2018ContextDependentDN, peyre2017weakly-supervised, zhang2017ppr-fcn, chen2019scene, peyre2019detecting} have provided solutions for zero-shot relationship learning. Some researchers recently proposed unbiased SGG\cite{yu2020cogtree, yang2018shuffle, tang2020unbiased, yan2020pcpl, wang2020tackling} to make the tail classes to receive more attention in a coarse-to-fine mode. 

The \bfseries third \mdseries challenge is that the visual appearance of the same relation varies greatly from scene to scene ({Fig.\ref{fig:Various_Examples}a and \ref{fig:Various_Examples}d}). This makes the feature extraction phase more challenging. As we have described in Section \ref{sec:Using_language_Priors}, a great deal of methods focuses on semantic features, trying to make up for the lack of visual features. However, we have emphasized that visual relationships are incidental and scene-specific. This requires us to think from the bottom up and try to extract more discriminative visual features.

\textcolor{black}{The \bfseries fourth \mdseries challenge is the lack of clarity/consensus in the definition of the relationships. It is always difficult to give mutually exclusive definitions to the predicate categories as opposed to objects, which have clear meaning. As a result, one relationship can be labelled with different but reasonable predicates, making the datasets noisy and the general SGG task ill-posed. Providing a well-defined relationship set is therefore one of the key challenges of the SGG task. }

\textcolor{black}{The \bfseries fifth \mdseries challenge is the evaluation metric. Even though many evaluation metrics are used to assess the performance of the proposed networks and Recall@K or meanRecall@K are common and widely adopted, none of them can provide perfect statistics on how well the model performs on the SGG task. When Recall@50 equals 100, does that mean that the model generates the perfect scene graph for an image? Of course not. The existing evaluation metrics only reveal the relative performance, especially in the current research stage. As the research on SGG progresses, the evaluation metrics and benchmark datasets will pose a great challenge.}

\subsection{Opportunities} 

The community has published hundreds of scene graph models and has obtained a wealth of research result. We think there are several avenues for future work. \textcolor{black}{Researchers will be motivated to explore more models as a result of the above challenges. Besides, on} the one hand, from the learning point of view, building a large dataset with fine-grained labels and accurate annotations is necessary and significant. It contains as many scenes as possible, preferably constructed by computer vision experts. The models trained on such a dataset will have better performance on visual semantic and develop a broader understanding of our visual world. However, this is a very challenging and expensive task. On the other hand, from the application point of view, we can design the models by subdividing the scene to reduce the imbalance of the relationship distribution. Obviously, the categories and probability distributions of visual relationships are different in different scenarios. Of course, even the types of objects are different. \textcolor{black}{As a result, we can design relationship detection models for different scenarios and employ ensemble learning methods to promote scene graph generation applications. }

\textcolor{black}{Another area of research is 3D scene graphs. An initial step is to define an effective and unified 3D scene graph structure, along with what information it should encode. A 2D image is a two-dimensional projection of a 3D world scene taken from a specific viewpoint. It is the specific viewpoint that makes some descriptions of spatial relationships in 2D images meaningful. Taking the triplet of $\left\langle women, is\  behind, fire\ hydrant\right\rangle$ in Fig. 1 as an example, the relation ``is behind" makes sense because of the viewpoint. But, how can a relation be defined as ``is behind" in 3D scenes without a given viewpoint? Therefore, the challenge is how to define such spatial semantic relationships in 3D scenes without simply resorting to 2.5D scenes (for example, RGB-D data captured from a specific viewpoint).} Armeni \emph{et al}. augment the basic scene graph structure with essential 3D information and generate a 3D scene graph which extends the scene graph to 3D space and ground semantic information there\cite{armeni20193d, kim20193-d}. However, their proposed structure representation does not have expansibility and generality. Second, because 3D information can be grounded in many storage formats, which are fragmented to specific types based on the visual modality (e.g., RGB-D, point clouds, 3D mesh/CAD models, etc.), the presentation and extraction of 3D semantic information has technological challenges.

\section{Conclusion}

This paper provides a comprehensive survey of the developments in the field of scene graph generation using deep learning techniques. We first introduced the representative works on 2D scene graph, spatio-temporal scene graph and 3D scene graph in different sections, respectively. Furthermore, we provided a summary of some of the most widely used datasets for visual relationship and scene graph generation, which are grouped into 2D images, video, and 3D representation, respectively. The performance of different approaches on different datasets are also compared. Finally, we discussed the challenges, problems and opportunities on the scene graph generation research. We believe this survey can promote more in-depth ideas used on SGG.


%




\ifCLASSOPTIONcaptionsoff
  \newpage
\fi

\end{document}